%% file: pryor-acl23.tex
\pdfoutput=1
\documentclass[11pt]{article}

\usepackage{ACL2023}

\usepackage{times}
\usepackage{latexsym}

\usepackage[T1]{fontenc}
\usepackage[utf8]{inputenc}

\usepackage{microtype}

\usepackage{inconsolata}

\title{Using Domain Knowledge to Guide Dialog \\ Structure Induction via Neural Probabilistic Soft Logic}

\usepackage[ruled,linesnumbered]{algorithm2e}
\SetKwInput{KwInput}{Input}
\usepackage{amsmath}
\usepackage{amssymb}
\usepackage{amsthm}
\usepackage{booktabs}
\usepackage{calc}
\usepackage{changepage}
\usepackage{graphicx}
\usepackage{multirow}
\usepackage{subcaption}
\usepackage{float}
\usepackage{comment}

\usepackage{macros}

\author{
    Connor Pryor \thanks{Work done during internship at Google.} \\
    UC Santa Cruz \\
    \texttt{cfpryor@ucsc.edu} \\
    \And
    Quan Yuan \\
    Google Research \\
    \texttt{yquan@google.com} \\
    \And
    Jeremiah Liu \\
    Google Research \\
    Harvard University \\
    \texttt{jereliu@google.com} \\
    \And
    Mehran Kazemi \\
    Google Research \\
    \texttt{mehrankazemi@}\\
    \texttt{google.com} \\
    \AND
    Deepak Ramachandran \\
    Google Research \\
    \texttt{ramachandrand@google.com} 
    \And
    Tania Bedrax-Weiss \\
    Google Research \\
    \texttt{tbedrax@google.com}
    \And
    Lise Getoor \\
    UC Santa Cruz \\
    \texttt{getoor@ucsc.edu} \\
}

\begin{document}
    \maketitle

    \begin{abstract}
        \input{sections/abstract.tex}
    \end{abstract}

    \input{sections/introduction.tex}

    \input{sections/related-work.tex}

    \input{sections/background.tex}

    \input{sections/method.tex}

    \input{sections/evaluation.tex}

    \input{sections/conclusion.tex}

    \input{sections/ack}

    \bibliography{pryor-acl23.bib}
    \bibliographystyle{acl_natbib}

    \begin{appendix}
        \input{appendix/constraints.tex}
        \input{appendix/datasets}
        \input{appendix/extended_experiments}
    \end{appendix}
\end{document}

%% file: sections/abstract.tex
\textit{Dialog Structure Induction} (DSI) is the task of inferring the latent dialog structure (i.e., a set of dialog states and their temporal transitions) of a given goal-oriented dialog.
It is a critical component for modern dialog system design and discourse analysis.
Existing DSI approaches are often purely data-driven, deploy models that infer latent states without access to domain knowledge, underperform when the training corpus is limited/noisy, or have difficulty when test dialogs exhibit distributional shifts from the training domain.
This work explores a neural-symbolic approach as a potential solution to these problems.
We introduce \textit{\longname} (\shortname), a principled approach that injects symbolic knowledge into the latent space of a generative neural model.
We conduct a thorough empirical investigation on the effect of \shortname{} learning on hidden representation quality, few-shot learning, and out-of-domain generalization performance.
Over three dialog structure induction datasets and across unsupervised and semi-supervised settings for standard and cross-domain generalization, the injection of symbolic knowledge using \shortname{} provides a consistent boost in performance over the canonical baselines.

%% file: sections/introduction.tex
\begin{figure}[t]
    \centering
    \includegraphics[width=0.49\textwidth]{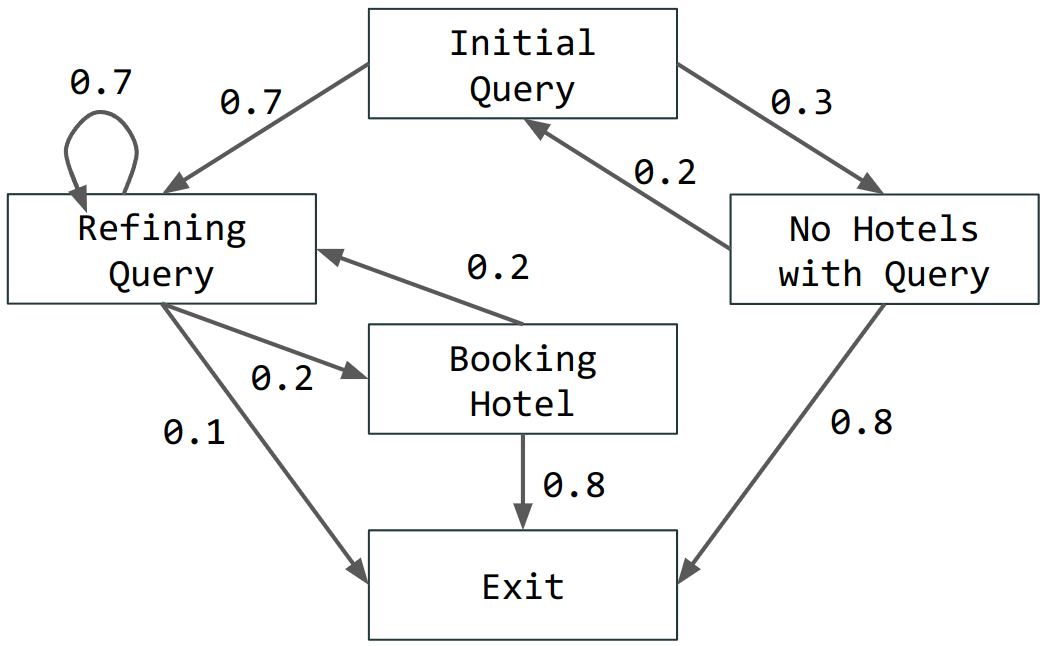}

    \caption{
        Example dialog structure for the goal-oriented task booking a hotel.
    }
    \label{fig:hotel}
\end{figure}

\section{Introduction}

The seamless integration of prior domain knowledge into the neural learning of language structure has been an open challenge in the machine learning and natural language processing communities.
In this work, we inject symbolic knowledge into the neural learning process of a two-party \textit{dialog structure induction} (DSI) task \citep{zhai:acl14, shi:acl19}.
This task aims to learn a graph, known as the \textit{dialog structure}, capturing the potential flow of states occurring in a dialog dataset for a specific task-oriented domain, e.g., \figref{fig:hotel} represents a possible dialog structure for the goal-oriented task of booking a hotel.
Nodes in the dialog structure represent conversational topics or \textit{dialog acts} that abstract the intent of individual utterances, and edges represent transitions between dialog acts over successive turns of the dialog.

Similar to the motivation described in \cite{shi:acl19}, previous work in DSI has been split between supervised and unsupervised methods.
In particular, traditional supervised methods \cite{jurafsky:ics97} relied on dialog structure hand-crafted by human domain experts.
Unfortunately, this process is labor-intensive and, in most situations, does not generalize easily to new domains.
Therefore, recent work attempts to overcome this limitation by studying unsupervised DSI; e.g., \textit{hidden Markov models} \cite{chotimongkol:thesis08,ritter:acl10,zhai:acl14} and more recently \textit{Variational Recurrent Neural Networks} (VRNN) \cite{chung:neurips15,shi:acl19}.
However, being purely data-driven, these approaches have difficulty with limited/noisy data and cannot easily exploit domain-specific or domain-independent constraints on dialog that may be readily provided by human experts (e.g., \textit{Greet} utterances are typically made in the first couple of turns).

In this work, we propose \textit{\longname} (\shortname).
This practical neuro-symbolic approach improves the quality of learned dialog structure by infusing domain knowledge into the end-to-end, gradient-based learning of a neural model.
We leverage \textit{Probabilistic Soft Logic} (PSL), a well-studied soft logic formalism, to express domain knowledge as soft rules in succinct and interpretable first-order logic statements that can be incorporated easily into differentiable learning
\citep{bach:jmlr17, pryor:ijcai23}.
This leads to a simple method for knowledge injection with minimal change to the SGD-based training pipeline of an existing neural generative model.

Our key contributions are:
1) We propose \shortname, which introduces a novel smooth relaxation of PSL constraints tailored to ensure a rich gradient signal during back-propagation;
2) We evaluate \shortname{} over synthetic and realistic dialog datasets under three settings: standard generalization, domain generalization, and domain adaptation.
We show quantitatively that injecting domain knowledge provides a boost over unsupervised and few-shot methods;
and 3) We comprehensively investigate the effect of soft logic-augmented learning on different aspects of the learned neural model by examining its quality in representation learning and structure induction.

%% file: sections/related-work.tex
\section{Related Work}

\textit{Dialog Structure Induction} (DSI) refers to the task of inferring latent states of a dialog without full supervision of the state labels.
Earlier work focus on building advanced clustering methods, e.g., topic models, HMM, GMM \citep{zhai:acl14}, which are later combined with pre-trained or task-specific neural representations \citep{nath2021tscan, lv2021task, qiu2022structure}.
Another line of work focuses on inferring latent states using neural generative models, most notably \textit{Direct-Discrete Variational Recurrent Neural Networks} (DD-VRNN) \citep{shi:acl19}, with later improvements including BERT encoder \citep{chen2021dsbert}, GNN-based latent-space model \citep{sun2021unsupervised,xu2021discovering}, structured-attention decoder \citep{qiu2020structured}, and database query modeling \citep{hudevcek2022learning}.
Finally, \citet{zhang2020probabilistic, wu2020improving} explored DSI in a semi-supervised and few-shot learning context.
No work has explored DSI with domain knowledge as weak supervision or conducted a comprehensive evaluation of model performance across different generalization settings (i.e., unsupervised, few-shot, domain generalization, and domain adaptation).

A related field of work, Neuro-Symbolic computing (NeSy), is an active area of research that aims to incorporate logic-based reasoning with neural computation.
This field contains a plethora of different neural symbolic methods and techniques.
The methods that closely relate to our line of work seek to enforce constraints on the output of a neural network \cite{hu2016harnessing, donadello:ijcai17, diligenti:icmla17, mehta:emnlp18, xu:icml18, nandwani:neurips19}.
For a more in-depth introduction, we refer the reader to these excellent recent surveys: \citenoun{besold:arxiv17} and \citenoun{deraedt:ijcai20}.
These methods, although powerful, are either: specific to the domain they work in, do not use the same soft logic formulation, have not been designed for unsupervised systems, or have not been used for dialog structure induction.

Finally, our method is most closely related to the novel NeSy approaches of \textit{Neural Probabilistic Soft Logic} (NeuPSL) \cite{pryor:ijcai23}, \textit{DeepProbLog} (DPL) \cite{manhaeve:ai21}, and \textit{Logic Tensor Networks} (LTNs) \cite{badreddine:ai22}.
LTNs instantiate a model which forwards neural network predictions into functions representing symbolic relations with real-valued or fuzzy logic semantics, and DeepProbLog uses the output of a neural network to specify probabilities of events.
The mathematical formulation of LTNs and DPL differs from our underlying soft logic distribution.
NeuPSL unites state-of-the-art symbolic reasoning with the low-level perception of deep neural networks through Probabilistic Soft Logic (PSL).
Our method uses a NeuPSL formulation; however, we introduce a novel variation to the soft logic formulation, develop theory for unsupervised tasks, introduce the whole system in Tensorflow, and apply it to dialog structure induction.

%% file: sections/background.tex
\section{Background}
\label{sec:background}

Our neuro-symbolic approach to dialog structure induction combines the principled formulation of probabilistic soft logic (PSL) rules with a neural generative model.
In this work, we use the state-of-the-art Direct-Discrete Variational Recurrent Neural Network (DD-VRNN) as the base model \citep{shi:acl19}.
We start by introducing the syntax and semantics for DD-VRNN and PSL.

\begin{figure*}[t]
    \centering
    \includegraphics[width=0.95\textwidth]{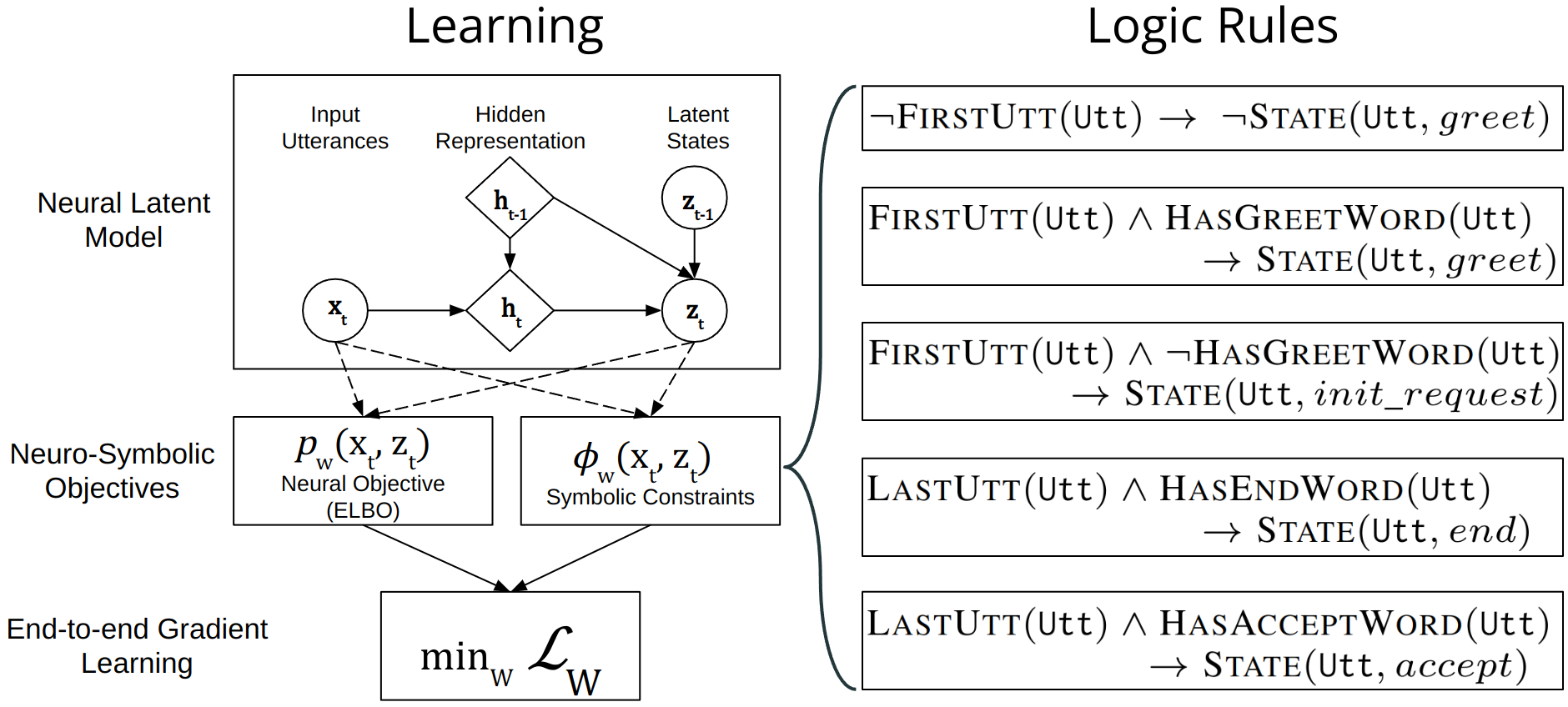}

    \caption{
        The high-level pipeline of the \shortname  $\;$ learning procedure.
    }
    \label{fig:pipeline}
\end{figure*}

\subsection{Direct Discrete Variational Recurrent Neural Networks}
\label{sec:background-dd-vrnn}

A Direct Discrete Variational Recurrent Neural Networks (DD-VRNN) \cite{shi:acl19} is an expansion to the popular Variational Recurrent Neural Networks (VRNN) \cite{chung:neurips15}, which constructs a sequence of VAEs and associates them with states of an RNN.
The main difference between the DD-VRNN and a traditional VRNN is the priors of the latent states $z_t$.
They directly model the influence of $z_{t-1}$ on $z_t$, which models the transitions between different latent (i.e., dialog) states.
To fit the prior into the variational inference framework, an approximation of $p(z_t | x_{<t}, z_{<t})$ is made which changes the distribution to $p(z_t | z_{t-1})$:
\begin{align}
    p(x_{\leq T}, z_{\leq T})
    & = \prod_{t=1}^T p(x_t | z_{\leq t}, x_{< t}) p(z_t | x_{<t}, z_{t-1})  
    \nonumber\\
    & \approx \prod_{t=1}^T p(x_t | z_{\leq t}, x_{< t}) p(z_t | z_{t-1})
\end{align}
$z_t$ is modeled as $ z_t \sim softmax(\phi_{\tau}^{prior}(z_{t-1}))$ for a feature extraction neural network for the prior $\phi_{\tau}^{prior} $.
Lastly, the objective function used in the DD-VRNN is a timestep-wise variational lower bound \cite{chung:neurips15}  augmented with a bag-of-word (BOW) loss and Batch Prior Regularization (BPR) \cite{zhao:acl17,zhao:acl18}, i.e.:
\begin{align}
        & \mathcal{L}_{VRNN} = \mathbb{E}_{q(z \leq T | x \leq T)}[\log p(x_t | z_{\leq t}, x_{<t}) + \nonumber \\
        & \sum_{t=1}^{T}-KL(q(z_t | x_{\leq t}, z_{<t}) || p(z_t | x_{<t}, z_{<t}))]
\end{align}
so that the full objective function is
\begin{align}
    \mathcal{L}_{DD-VRNN} =\mathcal{L}_{VRNN} + \lambda * \mathcal{L}_{bow}
    \label{eqn:background-dd-vrnn-loss}
\end{align}
where $\lambda$ is a tunable weight and $\mathcal{L}_{bow}$ is the BOW loss.
For further details on $\mathcal{L}_{bow}$ see \secref{sec:method_bow_reweighting} and \citenoun{shi:acl19}.
To expand this to a semi-supervised domain, the objective is augmented as:
\begin{align}
    & \mathcal{L}_{DD-VRNN} = \nonumber \\
    & \hspace{0.5cm} \mathcal{L}_{VRNN} + \lambda * \mathcal{L}_{bow} + \mathcal{L}_{supervised}
    \label{eqn:dd-vrnn-semi-supervised}
\end{align}
where $\mathcal{L}_{supervised}$ is the loss between the labels and predictions, e.g., \textit{cross-entropy}.

\subsection{Probabilistic Soft Logic}
\label{sec:probabilistic-soft-logic}

This work introduces soft constraints in a declarative fashion, similar to Probabilistic Soft Logic (PSL).
PSL is a declarative statistical relational learning (SRL) framework for defining a particular probabilistic graphical model, known as a \textit{hinge-loss Markov random field} (HL-MRF) \cite{bach:jmlr17}.
PSL models relational dependencies and structural constraints using first-order logical rules, referred to as \textit{templates}, with arguments known as \textit{atoms}.
For example, the statement ``the first utterance in a dialog is likely to belong to the \textit{greet} state" can be expressed as:
\begin{align}
    % w: 
    \pslpred{FirstUtt}(\pslarg{U}) \rightarrow \pslpred{State}(\pslarg{U}, greet)
    \label{eqn:psl-rule-dialog-example}
\end{align}
Where ($\pslpred{FirstUtt}(\pslarg{U})$, $\pslpred{State}(\pslarg{U}, greet)$) are the \textit{atoms} (i.e., atomic boolean statements) indicating, respectively, whether an utterance $\pslarg{U}$ is the first utterance of the dialog, or if it belongs to the state \texttt{greet}.
The atoms in a PSL rule are \emph{grounded} by replacing the free variables (such as $\pslarg{U}$ above) with concrete instances from a domain of interest (e.g., the concrete utterance 'Hello!'); we call these the 
\emph{grounded atoms}.
The observed variables and target/decision variables of the probabilistic model correspond to ground atoms constructed from the domain, e.g., $\pslpred{FirstUtt}('Hello!')$ may %could 
be an observed variable and $\pslpred{State}('Hello!', greet)$ may be a target variable.

PSL performs inference over soft logic constraints by allowing the originally Boolean-valued atoms to take continuous truth values in the interval $[0,1]$.
Using this relaxation, PSL replaces logical operations with a form of soft logic called \textit{Lukasiewicz} logic \cite{klir:book95}:
\begin{align}
    A \psland B &= max(0.0, A \pslsum B - 1.0) \nonumber \\
    A \pslor B &= min(1.0, A \pslsum B) \nonumber \\
    \pslneg A &= 1.0 - A
\end{align}
where $A$ and $B$ represent either ground atoms or logical expressions over atoms and take values in $[0,1]$.
For example, PSL will convert the statement from \eqnref{eqn:psl-rule-dialog-example}, into the following:
\begin{align}
    min\{1,~& 1 - \pslpred{FirstUtt}(\pslarg{U}) \pslsum \pslpred{State}(\pslarg{U}, greet))\}
    \label{eqn:constraint-example}
\end{align}
since $A \rightarrow B \equiv \pslneg A \pslor B$. 
In this way, we can create a collection of functions $\{\ell_{i}\}_{i=1}^m$, called \textit{templates}, that map data to $[0, 1]$.
Using the templates, PSL defines a conditional probability density function over the unobserved random variables $\mathbf{y}$ given the observed data $\mathbf{x}$ known as the \textit{Hinge-Loss Markov Random Field} (HL-MRF):
\begin{align}
    P(\mathbf{y}|\mathbf{x}) \propto exp(-\sum_{i=1}^{m} \lambda_{i} \cdot \phi_{i}(\mathbf{y}, \mathbf{x}))
    \label{eqn:psl-hl-mrf}
\end{align}
Here $\lambda_{i}$ is a non-negative weight and $\phi_{i}$ a \textit{potential function} based on the templates:
\begin{align}
    \phi_{i}(\mathbf{y}, \mathbf{x}) = max\{0, \ell_{i}(\mathbf{y}, \mathbf{x})\}
\end{align}
Then, inference for the model predictions $\mathbf{y}$ proceeds by Maximum A Posterior (MAP) estimation, i.e., by maximizing the objective function $P(\mathbf{y}| \mathbf{x})$ (eq. \ref{eqn:psl-hl-mrf}) with respect to $\mathbf{y}$.

%% file: sections/method.tex
\section{\longname}
\label{sec:method}

In this section, we describe our approach to integrating domain knowledge and neural network-based 
dialog structure induction.
Our approach integrates an unsupervised neural generative model with dialog rules using soft constraints.
We refer to our approach as \textit{\longname} \ (\shortname).
In the following, we define the dialog structure learning problem, describe how to integrate the neural and symbolic losses, and highlight essential model components that address optimization and representation-learning challenges under gradient-based neuro-symbolic learning.

\paragraph{Problem Formulation}
Given a goal-oriented dialog corpus $\mathcal{D}$, we consider the DSI problem of learning a graph $G$ underlying the corpus.
More formally, a \emph{dialog structure} is defined as a directed graph $G=(S, P)$, where $S=\{s_1, \dots, s_m\}$ encodes a set of dialog states, and $P$ a probability distribution $p(s_t|s_{<t})$ representing the likelihood of transition between states (see \figref{fig:hotel} for an example).
Given the underlying dialog structure $G$, a dialog $d_i=\{x_1, \dots, x_T\} \in \mathcal{D}$ is a temporally-ordered set of utterances $x_t$.
Assume $x_t$ is defined according to an utterance distribution conditional on past history $p(x_t|s_{\leq t}, x_{<t})$, and the state $s_t$ is defined according to $p(s_t|s_{< t})$.
Given a dialog corpus $\mathcal{D}=\{d_i\}_{i=1}^n$, the task of DSI is to learn a directed graphical model $G=(S, P)$ as close to the underlying graph as possible.

\subsection{Integrating Neural and Symbolic Learning under \shortname}

We now introduce how the \shortname{} approach formally integrates the DD-VRNN with the soft symbolic constraints to allow for end-to-end gradient training.
To begin, we define the relaxation of the symbolic constraints to be the same as described in \secref{sec:probabilistic-soft-logic}.
With this relaxation, we can build upon the foundations developed by \citenoun{pryor:ijcai23} on Neural Probabilistic Soft Logic (NeuPSL) by augmenting the standard unsupervised DD-VRNN loss with a constraint loss.
\figref{fig:pipeline} provides a graphical representation of this integration of the DD-VRNN and the symbolic constraints.
Intuitively, \shortname \ can be described in three parts: instantiation, inference, and learning.

Instantiation of a \shortname \ model uses a set of first-order logic templates to create a set of potentials that define a loss used for learning and evaluation.
Let $p_{\textbf{w}}$ be the DD-VRNN's predictive function of latent states with hidden parameters $\mathbf{w}$ and input utterances $\mathbf{x}_{vrnn}$.
The output of this function, defined as $p_{\textbf{w}}(\mathbf{x}_{vrnn})$, will be the probability distribution representing the likelihood of each latent class for a given utterance.
Given a first-order symbolic rule $\ell_i(\mathbf{y}, \mathbf{x})$, where $\mathbf{y}=p_{\textbf{w}}(\mathbf{x}_{vrnn})$ is the decision variable representing the latent state prediction from the neural model, and $\mathbf{x}$ represents the symbolic observations not incorporated in the neural features $\mathbf{x}_{vrnn}$, \shortname{} instantiates a set of \textbf{deep hinge-loss potentials} of the form:
\begin{align}
    \phi_{\textbf{w}, i}(\mathbf{x}_{vrnn}, \mathbf{x})
    = \max(0, \ell_i(p_{\textbf{w}}(\mathbf{x}_{vrnn}), \mathbf{x}))
\end{align}
For example, in reference to \eqnref{eqn:constraint-example}, the decision variable $\mathbf{y}=p_{\textbf{w}}(\mathbf{x}_{vrnn})$ is associated with the $\pslpred{State}(\mathbf{x}, greet)$ random variables, leading to:
\begin{align}
    \ell_i&(p_{\textbf{w}}(\mathbf{x}_{vrnn}), \mathbf{x}) = \nonumber \\
    & min\{1, 1 - \pslpred{FirstUtt}(\pslarg{U}) + p_{\textbf{w}}(\mathbf{x}_{vrnn})\}
\end{align}
With the instantiated model described above, the \shortname \ inference objective is broken into a \textit{neural inference} objective and a \textit{symbolic inference} objective.
The neural inference objective is computed by evaluating the DD-VRNN model predictions with respect to the standard loss function for DSI. Given the deep hinge-loss potentials $\{\phi_{\textbf{w}, i}\}_{i=1}^m$, the symbolic inference objective is the HL-MRF likelihood (Equation \ref{eqn:psl-hl-mrf}) evaluated at the decision variables $\mathbf{y}=p_{\textbf{w}}(\mathbf{x}_{vrnn})$:
\begin{align}
    P_{\textbf{w}}(\mathbf{y}|&\mathbf{x}_{vrnn}, \mathbf{x}, \mathbf{\lambda}) = \\ \nonumber
    & exp\big( -\sum_{i=1}^m \lambda_{i} \cdot \phi_{\textbf{w}, i}(\mathbf{x}_{vrnn}, \mathbf{x}) \big)
    \label{eq:deep_hl_mrf}
\end{align}
Under the \shortname, the decision variables $\mathbf{y}=p_{\textbf{w}}(\mathbf{x}_{vrnn})$ are implicitly controlled by neural network weights $\textbf{w}$, therefore the conventional MAP inference in symbolic learning for decision variables $\mathbf{y}^*=\argmin_{\mathbf{y}} P(\mathbf{y}| \mathbf{x}_{vrnn}, \mathbf{x}, \mathbf{\lambda})$ can be done simply via neural weight minimization $\argmin_{\mathbf{w}} P_{\textbf{w}}(\mathbf{y}|\mathbf{x}_{vrnn}, \mathbf{x}, \mathbf{\lambda})$.
As a result, \shortname{} learning minimizes a constrained optimization objective:
\begin{align}
    \mathbf{w}^{*} = \argmin_{\mathbf{w}}
    \Big[ 
    \mathcal{L}_{DD-VRNN} + 
    \mathcal{L}_{constraint}
    \Big]
\end{align}
where we define the constraint loss to be the log-likelihood of the HL-MRF distribution (\ref{eq:deep_hl_mrf}):
\begin{align}
    & \mathcal{L}_{Constraint} = - log P_\textbf{w}(\mathbf{y}|\mathbf{x}_{vrnn}, \mathbf{x}, \mathbf{\lambda}).
\end{align}

\subsection{Improving soft logic constraints for gradient learning}
\label{sec:log-constraints}

The straightforward linear soft constraints used by the classic Lukasiewicz relaxation fail to pass back gradients with a magnitude and instead pass back a direction (e.g., $\pm 1$).
Formally, the gradient of a potential $\phi_{\textbf{w}}(\mathbf{x}_{vrnn}, \mathbf{x})
= \max(0, \ell(p_{\textbf{w}}(\mathbf{x}_{vrnn}), \mathbf{x}))$ with respect to $\textbf{w}$ is:
\begin{align}
    \frac{\partial}{\partial \textbf{w}}  \phi_{\textbf{w}} & = \frac{\partial}{\partial \textbf{w}} \ell(p_{\textbf{w}}, \mathbf{x}) \cdot 1_{\phi_{\textbf{w}} > 0} \nonumber\\
    & = \Big[ \frac{\partial}{\partial p_{\mathbf{w}}} \ell(p_{\textbf{w}}, \mathbf{x}) \Big] \cdot  \frac{\partial}{\partial \textbf{w}} p_{\textbf{w}}
    \cdot
    1_{\phi_{\textbf{w}} > 0}
\end{align}
Here $\ell(p_{\textbf{w}}, \mathbf{x}) = a \cdot p_{\textbf{w}}(\mathbf{x}_{vrnn}) + b$ where $a, b \in \mathbb{R}$ and $p_{\textbf{w}}(\mathbf{x}_{vrnn}) \in [0,1]$, which leads to
the gradient $\frac{\partial}{\partial p_{\mathbf{w}}} \ell(p_{\textbf{w}}, \mathbf{x})=a$.
Observing the three Lukasiewicz operations described in \secref{sec:probabilistic-soft-logic}, it is clear that $a$ will always result in $\pm 1$ unless there are multiple $p_{\textbf{w}}(\mathbf{x}_{vrnn})$ per constraint.

As a result, this classic soft relaxation leads to a naive, non-smooth gradient:
\begin{align}
\frac{\partial}{\partial \textbf{w}} \phi_\textbf{w} = \Big[a 1_{\phi_{\textbf{w}} > 0}\Big] \cdot \frac{\partial}{\partial \textbf{w}} p_{\textbf{w}}
    \label{eqn:potential-sub-gradient}
\end{align}
that is mostly consists of the predictive probability gradient $\frac{\partial}{\partial \textbf{w}} p_{\textbf{w}}$. It barely informs the model of the degree to which $p_\textbf{w}$ satisfies the symbolic constraint $\phi_\textbf{w}$ (other than the non-smooth step function $1_{\phi_{\textbf{w}} > 0}$), thereby creating challenges in gradient-based learning.

In this work, we propose a novel log-based relaxation that provides smoother and more informative gradient information for the symbolic constraints:
\begin{align}
    \psi_\textbf{w}(\mathbf{x}) & = \log \big(\phi_\textbf{w}(\mathbf{x}_{vrnn}, \mathbf{x})\big) \\ \nonumber
    & = \hspace{0.02cm} \log\big(\max(0, \ell(p_{\textbf{w}}(\mathbf{x}_{vrnn}), \mathbf{x}) ) \big)
\end{align}
This seemingly simple transformation brings a non-trivial change to the gradient behavior:
\begin{align}
    \frac{\partial}{\partial \textbf{w}} \psi_{\textbf{w}} & = \frac{1}{\phi_\textbf{w}} \cdot \frac{\partial}{\partial \textbf{w}}  \phi_{\textbf{w}} \\ \nonumber
    & = \Big[ \frac{a}{\phi_\textbf{w}} 1_{\phi_{\textbf{w}} > 0} \Big] \cdot \frac{\partial}{\partial \textbf{w}}  p_{\textbf{w}}
\end{align}
As shown, the gradient from the symbolic constraint now contains a new term $\frac{1}{\phi_\textbf{w}}$.
It informs the model of the degree to which the model prediction satisfies the symbolic constraint $\ell$ so that it is no longer a discrete step function with respect to $\phi_\textbf{w}$.
As a result, when the satisfaction of a rule $\phi_\textbf{w}$ is non-negative but low (i.e., uncertain), the gradient magnitude will be high, and when the satisfaction of the rule is high, the gradient magnitude will be low. In this way, the gradient of the symbolic constraint terms $\phi_i$ now guides the neural model to more efficiently focus on learning the challenging examples that don't obey the existing symbolic rules. This leads to more effective collaboration between the neural and the symbolic components during model learning and empirically leads to improved generalization performance (\secref{sec:experimental_evaluation}).

\subsection{Stronger control of posterior collapse via weighted bag of words}
\label{sec:method_bow_reweighting}

It is essential to avoid a collapsed VRNN solution, where the model puts all of its predictions in just a handful of states.
This problem has been referred to as the vanishing latent variable problem \cite{zhao:acl17}.
\citenoun{zhao:acl17} address this by introducing a \textit{bag-of-word (BOW) loss} to VRNN modeling which requires a network to predict the bag-of-words in response $x$.
They separate $x$ into two variables: $x_o$ (word order) and $x_{bow}$ (no word order), with the assumption that they are conditionally independent given $z$ and $c$:
\begin{align}
    p(x,z|c) = p(x_o|z,c)p(x_{bow}|z,c)p(z|c).
\end{align}
Here, $c$ is the dialog history: the preceding utterances, conversational floor (1 if the utterance is from the same speaker and 0 otherwise), and meta-features (e.g., the topic).
Let $f$ be the output of a multilayer perception with parameters $z,x$, where $f \in \mathbb{R}^{V}$ with $V$ the vocabulary size.
Then the BOW probability is defined as $\log p(x_{bow}|z,c) = \log \prod_{t=1}^{|x|} \dfrac{e^{f_{x_t}}}{\sum_{j}^Ve^{f_j}}$, where $|x|$ is the length of $x$ and $x_t$ is the word index of the $t_{th}$ word in $x$.

To impose robust regularization against the posterior collapse, we use a tf-idf-based re-weighting scheme using the tf-idf weights computed from the training corpus.
Intuitively, this re-weighting scheme helps the model focus on reconstructing non-generic terms that are unique to each dialog state, which encourages the model to ``pull" the sentences from different dialog states further apart in its representations space to minimize the weighted BOW loss better.
In comparison, a model under the uniformly-weighted BOW loss may be distracted by reconstructing the high-prevalence terms (e.g., "what is," "can I," and "when") that are shared by all dialog states.
As a result, we specify the tf-idf weighted BOW probability as:
\begin{align}
    \log p(x_{bow}|z,c) = \log \prod_{t=1}^{|x|} \dfrac{w_{x_{i}}e^{f_{x_t}}}{\sum_{j}^Ve^{f_j}},
\end{align}
where $w_{x_t} = \dfrac{(1 -\alpha)}{N} + \alpha w_{x_t}'$, $N$ is the corpus size, $w_{x_t}'$ is the tf-idf word weight for the $x_{t}$ index, and $\alpha$ is a hyperparameter.
In \secref{sec:experimental_evaluation}, we explore how this alteration affects the performance and observe if the PSL constraints still provide a boost.

%% file: sections/evaluation.tex
\begin{table*}[t]
    \centering
    \resizebox{\textwidth}{!}{
        \begin{tabular}{ccc||cc|c}
            \toprule
                \multirow{3}{*}{Dataset} & \multirow{3}{*}{Setting} & \multirow{3}{*}{Method} & \multicolumn{2}{c}{Hidden Representation Learning} & \multirow{2}{*}{Structure Induction}\\
                & & & Full & Few-Shot & \\
                & & & ( Class-Balanced Accuracy ) & ( Class-Balanced Accuracy )  & ( AMI )\\
             \midrule
                \multirow{2}{*}{MultiWoZ} & \multirow{2}{*}{\shortstack{Standard\\Generalization}} & DD-VRNN & \textbf{0.804 ± 0.037} & 0.643 ± 0.038 & 0.451 ± 0.042\\
                & & \shortname{} & \textbf{0.806 ± 0.051} & \textbf{0.689 ± 0.038} & \textbf{0.618 ± 0.028} \\
            \midrule
                \multirow{2}{*}{\shortstack{SGD\\Synthetic}} & \multirow{2}{*}{\shortstack{Standard\\Generalization}} & DD-VRNN & \textbf{0.949 ± 0.005} & 0.598 ± 0.019 & 0.553 ± 0.017 \\
                & & \shortname{} & 0.941 ± 0.009 & \textbf{0.765 ± 0.012} & \textbf{0.826 ± 0.006} \\
            \midrule
                \multirow{6}{*}{\shortstack{SGD\\Real}} & \multirow{2}{*}{\shortstack{Standard\\Generalization}} & DD-VRNN & \textbf{0.661 ± 0.015} & 0.357 ± 0.015 & 0.448 ± 0.019 \\
                &  & \shortname{} & \textbf{0.663 ± 0.015} & \textbf{0.517 ± 0.021} & \textbf{0.539 ± 0.048} \\
            \cmidrule{2-6}
                & \multirow{2}{*}{\shortstack{Domain\\Generalization}} & DD-VRNN & 0.268 ± 0.012 & 0.320 ± 0.029 & 0.476 ± 0.029 \\
                & & \shortname{} & \textbf{0.299 ± 0.009} & \textbf{0.528 ± 0.026} & \textbf{0.541 ± 0.036} \\
            \cmidrule{2-6}
                & \multirow{2}{*}{\shortstack{Domain\\Adaptation}} & DD-VRNN & \textbf{0.308 ± 0.011} & 0.505 ± 0.015 & 0.514 ± 0.028 \\
                & & \shortname{} & \textbf{0.297 ± 0.025} & \textbf{0.541 ± 0.023} & \textbf{0.559 ± 0.045} \\
            \bottomrule
        \end{tabular}
    }
    \caption{
        Test set performance on all datasets. All reported results are averaged over 10 splits. The highest-performing methods per dataset and learning setting are bolded. A random baseline has AMI zero and class-balanced accuracy equal to inverse class size (all less than 10\%, see Appendix Tables \ref{tab:ablation-multiwoz-synthetic}, \ref{tab:ablation-sgd-synthetic}, \ref{tab:ablation-multiwoz-ltn}).
    }
    \label{tab:structure-learning-performance}
\end{table*}

\begin{figure*}[!ht]
    \centering
    
    \includegraphics[width=\textwidth]{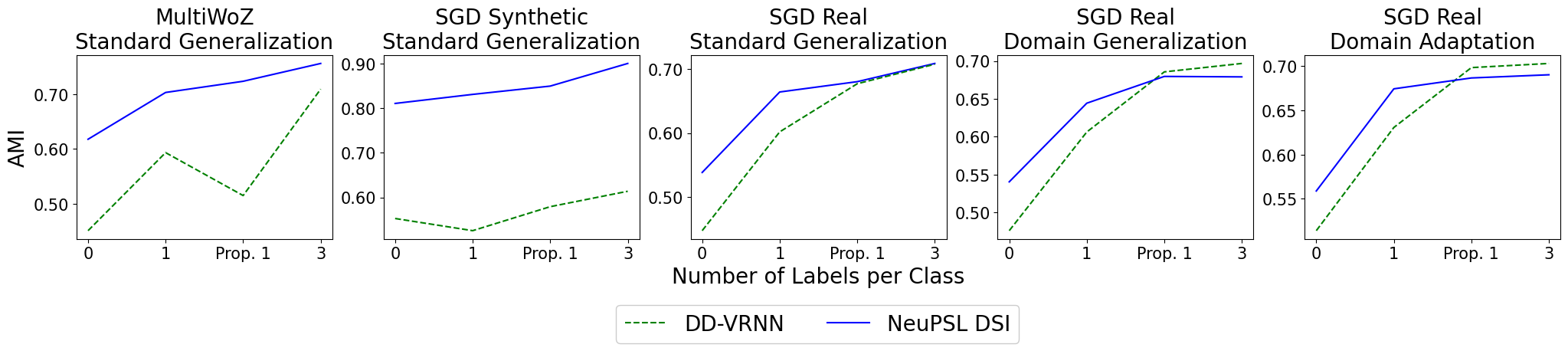}
    
    \caption{
        Average AMI for MultiWoZ, SGD Synthetic, and SGD Real (Standard Generalization, Domain Generalization, and Domain Adaptation) on three constrained few-shot settings: 1-shot, proportional 1-shot, and 3-shot.
        Hidden representation learning graphs are included in the Appendix.
    }
    \label{fig:main_results_few_shot}  
\end{figure*}

\section{Experimental Evaluation}
\label{sec:experimental_evaluation}

We evaluate the performance of \shortname{} on three task-oriented dialog corpuses in both unsupervised and highly constrained semi-supervised settings.
Further, we provide an extensive ablation on different aspects of the learned neural model.
We investigate the following questions:
Q1) How does \shortname{} perform in an unsupervised setting when soft constraints are incorporated into the loss?
Q2) When introducing few-shot labels to DD-VRNN training, do soft constraints provide a boost?
Q3) How do design choices such as log relaxation and re-weighted bag-of-words loss (introduced in \secref{sec:log-constraints}-\ref{sec:method_bow_reweighting}) impact performance?

\paragraph{Datasets} These questions are explored using three goal-oriented dialog datasets: MultiWoZ 2.1 synthetic \cite{campagna:acl20} and two versions of the Schema Guided Dialog (SGD) dataset; \textit{SGD-synthetic} (where the utterance is generated by a template-based dialog simulator) and \textit{SGD-real} (which replaces the machine-generated utterances of SGD-synthetic with its human-paraphrased counterparts) \citep{rastogi2020towards}.
For the SGD-real dataset, we evaluate over three unique data settings, \textit{standard generalization} (train and test over the same domain), \textit{domain generalization} (train and test over different domains), and \textit{domain adaptation}
(train on (potentially labeled) data from the training domain and unlabeled data from the test domain, and test on evaluation data from the test domain).
\appref{sec:data_app} describes further details.

\paragraph{Constraints} In the synthetic MultiWoZ setting, we introduce a set of 11 structural domain agnostic dialog rules.
An example of one of these rules can be seen in \eqnref{eqn:psl-rule-dialog-example}.
These rules are introduced to represent general facts about dialogs, with the goal of showing how the incorporation of a few expert-designed rules can drastically improve generalization performance.
For SGD settings, we introduce a single dialog rule that encodes the concept that dialog acts should contain utterances with correlated tokens, e.g., utterances containing 'hello' are likely to belong to the greet state.
This rule is designed to show the potential boost in performance a model can achieve from a simple source of prior information.
\appref{sec:data_app} contains further details.

\paragraph{Metrics and Methodology}
\label{sec:metrics_and_methodology}
The experimental evaluation examines two aspects: \textit{correctness of the learned latent dialog structure} and \textit{quality of the learned hidden representation}.

\textbf{Structure Induction.} 
To evaluate the model's ability to correctly learn the latent dialog structure, we adapt the Adjusted Mutual Information (AMI) metric from clustering literature (see \appref{sec:ami_app} for details).
AMI allows for a comparison between ground truth labels\footnote{These labels were only used for final evaluation, not for training or hyperparameter tuning.} (e.g., "greet", "initial request", etc.) and latent state predictions (e.g., $State_{1}, \cdot \cdot \cdot, State_{k}$).

\textbf{Representation Learning.}
A standard technique for evaluating the quality of unsupervised representation is \textit{linear probing}, i.e., train a lightweight linear \emph{probing} model on top of the frozen learned representation, and evaluate the linear model's generalization performance for supervised tasks \cite{tenney19}.
To evaluate the quality of the learned DD-VRNN, we train a supervised linear classifier on top of input features extracted from the penultimate layer of the DD-VRNN.
We evaluate with both full supervision and few-shot supervision.
Full supervision averages the class-balanced accuracy of two separate models that classify dialog acts (e.g., "greet", "initial request", etc.) and domains ("hotel", "restaurant", etc.) respectively.
Few-shot averages the class-balanced accuracy of models classifying dialog acts with 1-shot, 5-shot, and 10-shot settings.

\begin{table*}[t]
    \centering
    \resizebox{\textwidth}{!}{
        \begin{tabular}{c|ccc||cc|c}
            \toprule
                \multirow{3}{*}{Setting}&\multirow{3}{*}{\shortstack{Bag-of-Words\\ Weights}} & \multirow{3}{*}{\shortstack{Constraint\\ Loss}} & \multirow{3}{*}{Embedding} & \multicolumn{2}{c}{Hidden Representation Learning} & \multirow{2}{*}{Structure Induction}\\
                & & & & Full & Few-Shot & \\
                & & & & ( Class Balanced Accuracy ) & ( Class Balanced Accuracy )  & ( AMI )\\
             \midrule
                \multirow{8}{*}{\shortstack{Standard\\Generalization}} & Uniform & Linear & Bert & 0.588 ± 0.016 & \textbf{0.517 ± 0.021} & \textbf{0.539 ± 0.048} \\
                & Uniform & Linear & GloVe & 0.620 ± 0.023 & 0.428 ± 0.021 & 0.458 ± 0.024 \\
                & Uniform & Log & Bert & 0.600 ± 0.022 & \textbf{0.517 ± 0.023} & 0.520 ± 0.033 \\
                & Uniform & Log & GloVe & \textbf{0.650 ± 0.011} & 0.456 ± 0.014 & 0.532 ± 0.009 \\
                & tf-idf & Linear & Bert & 0.573 ± 0.022 & \textbf{0.521 ± 0.018} & 0.522 ± 0.024 \\
                & tf-idf & Linear & GloVe & 0.595 ± 0.014 & 0.379 ± 0.015 & 0.533 ± 0.048 \\
                & tf-idf & Log & Bert & 0.578 ± 0.021 & \textbf{0.510 ± 0.022} & 0.507 ± 0.060 \\
                & tf-idf & Log & GloVe & \textbf{0.653 ± 0.014} & 0.460 ± 0.009 & 0.534 ± 0.033 \\
            \midrule
                \multirow{8}{*}{\shortstack{Domain\\Generalization}} & Uniform & Linear & Bert & \textbf{0.597 ± 0.018} & \textbf{0.528 ± 0.026} & \textbf{0.541 ± 0.036} \\
                & Uniform & Linear & GloVe & \textbf{0.597 ± 0.012} & 0.391 ± 0.018 & 0.441 ± 0.030 \\
                & Uniform & Log & Bert & \textbf{0.598 ± 0.032} & \textbf{0.512 ± 0.021} & \textbf{0.517 ± 0.036} \\
                & Uniform & Log & GloVe & \textbf{0.608 ± 0.014} & 0.438 ± 0.017 & \textbf{0.508 ± 0.006} \\
                & tf-idf & Linear & Bert & 0.536 ± 0.026 & \textbf{0.518 ± 0.034} & \textbf{0.511 ± 0.018} \\
                & tf-idf & Linear & GloVe & 0.579 ± 0.033 & 0.360 ± 0.016 & 0.486 ± 0.057 \\
                & tf-idf & Log & Bert & 0.573 ± 0.018 & \textbf{0.516 ± 0.035} & 0.501 ± 0.064 \\
                & tf-idf & Log & GloVe & \textbf{0.599 ± 0.025} & 0.430 ± 0.020 & \textbf{0.505 ± 0.005} \\
            \midrule
                \multirow{8}{*}{\shortstack{Domain\\Adaptation}} & Uniform & Linear & Bert & 0.554 ± 0.135 & 0.492 ± 0.124 & \textbf{0.538 ± 0.107} \\
                & Uniform & Linear & GloVe & \textbf{0.667 ± 0.022} & \textbf{0.547 ± 0.025} & 0.419 ± 0.073 \\
                & Uniform & Log & Bert & 0.593 ± 0.049 & \textbf{0.541 ± 0.023} & \textbf{0.559 ± 0.045} \\
                & Uniform & Log & GloVe & 0.638 ± 0.024 & \textbf{0.555 ± 0.022} & 0.511 ± 0.045 \\
                & tf-idf & Linear & Bert & 0.584 ± 0.035 & \textbf{0.546 ± 0.023} & 0.494 ± 0.033 \\
                & tf-idf & Linear & GloVe & 0.593 ± 0.039 & 0.529 ± 0.022 & 0.463 ± 0.041 \\
                & tf-idf & Log & Bert & 0.597 ± 0.034 & \textbf{0.554 ± 0.025} & \textbf{0.549 ± 0.038} \\
                & tf-idf & Log & GloVe & 0.583 ± 0.029 & \textbf{0.534 ± 0.027} & 0.451 ± 0.044 \\
            \bottomrule
        \end{tabular}
    }
    \caption{
        Average performance for SGD real (Standard Generalization, Domain Generalization, and Domain Adaptation) over eight model settings (uniform/tf-idf bag-of-words weights, linear/log constraint loss, and BERT/GloVe embedding).
        The highest-performing settings are highlighted in bold.
    }
    \label{tab:ablation-sgd-standard}
\end{table*}

\subsection{Main Results}
\label{sec:main_results}

\tabref{tab:structure-learning-performance} summarizes the results of \shortname{} and \textit{DD-VRNN} in strictly unsupervised settings.
\shortname{} outperforms the strictly data-driven DD-VRNN on AMI by 4\%-27\% depending on the setting while maintaining or improving the hidden representation quality.
To reiterate, this improvement is achieved without supervision in the form of labels, but rather a few selected structural constraints.
Comparing AMI performance on SGD-real across different settings (standard generalization v.s. domain generalization/adaptation), we see the \shortname{} consistently improves over DD-VRNN, albeit with the advantage slightly diminished in the non-standard generalization settings.

To further understand how the constraints affect the model, we examine three highly constrained few-shot settings (1-shot, 3-shot, and proportional 1-shot) trained using the loss described in \eqnref{eqn:dd-vrnn-semi-supervised}.
The 1-shot and 3-shot settings are given one and three labels per class, while proportional 1-shot is provided the same number of labels as 1-shot with the distribution of labels proportional to the class size (classes below 1\% are not provided labels).
The results in \figref{fig:main_results_few_shot} show that in all settings, the introduction of labels improves performance.
This demonstrates that the soft constraints do not overpower learning but enable a trade-off between generalizing to priors and learning over labels.
In the SGD settings, however, as the number of labels increases, the pure data-driven approach performs as well or better than \shortname.

\begin{figure*}[t]
    \centering
    \includegraphics[width=\textwidth]{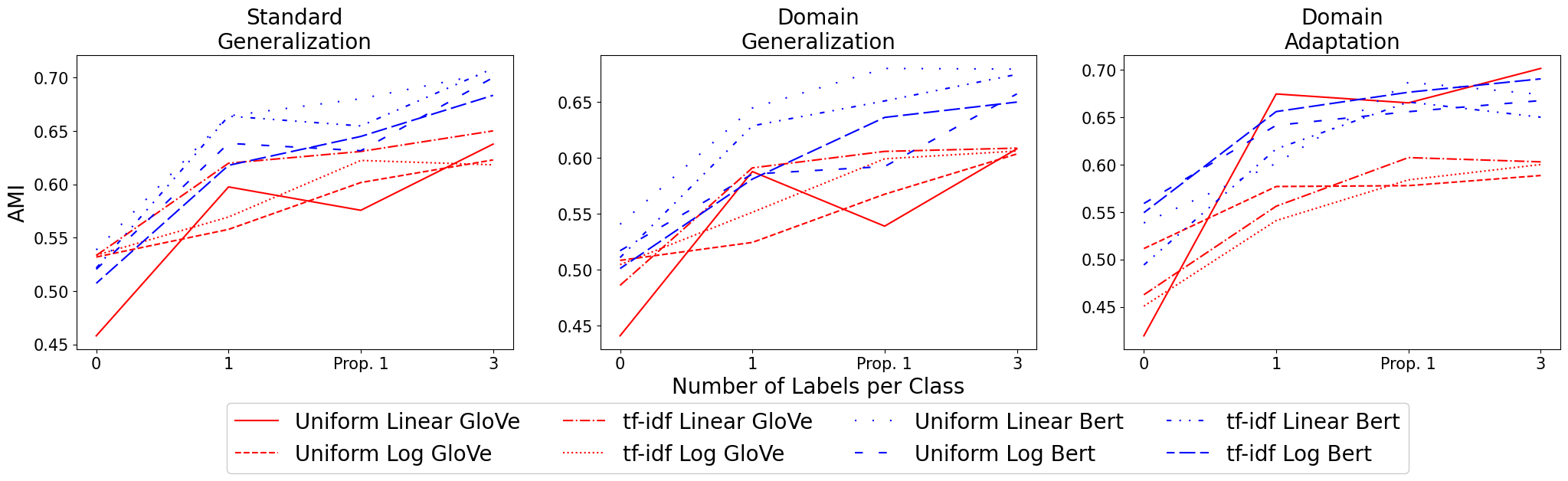}
    \caption{
        Average AMI performance for SGD Real (Standard Generalization, Domain Generalization, and Domain Adaptation) 
        on three highly constrained few-shot settings: 1 shot, proportional 1 shot, and 3 shot.
    }
    \label{fig:ablation_results_few_shot}
\end{figure*}

\subsection{Ablation Study}
\label{sec:ablation_study}
We provide an ablation on the SGD real dataset over three major method axes: parameterization of the constraint loss (linear v.s. log constraint loss, Section \ref{sec:log-constraints}), weighting scheme for the bag-of-words loss (uniform v.s. tf-idf weights, Section \ref{sec:method_bow_reweighting}), and the choice of underlying utterance embedding (BERT \cite{devlin-etal-2019-bert} v.s. GloVe \cite{pennington:emnlp14}) leading to a total of $2^3=8$ settings (Appendix \ref{sec:ablation_app} presents a further analysis for the MultiWoZ and SGD Synthetic datasets).
\tabref{tab:ablation-sgd-standard} summarizes the results for the SGD data set.
Highlighted in bold are the highest-performing setting or methods within a standard deviation of the highest-performing setting.
For structure induction, using a BERT embedding and uniform bag-of-words-weights generally produces the best AMI performance, while there is no significant difference between linear and log constraints.
However, when examining the hidden representation it is clear that the log relaxation outperforms or performs as well as its linear counterpart.
Additionally, \figref{fig:ablation_results_few_shot} summarizes the few-shot training results for the SGD data settings when training with 1-shot, proportional 1-shot, and 3-shots.
We see three methods generally on top in performance: uniform-log-bert, tf-idf-linear-bert, and uniform-linear-bert.
There seems to be no clear winner between uniform/tf-idf and linear/log; however, all three of these settings use BERT embeddings.

%% file: sections/conclusion.tex
\section{Discussion and Conclusions}

This paper introduces \shortname{}, a novel neuro-symbolic learning framework that uses differentiable symbolic knowledge to guide latent dialog structure learning.
Through extensive empirical evaluations, we illustrate how the injection of just a few domain knowledge rules 
significantly improves both correctness and hidden representation quality in this challenging unsupervised NLP task.

While \shortname{} sees outstanding success in the unsupervised settings, the introduction of additional labels highlights a potential limitation of \shortname{}.
If the domain knowledge introduced is weak or noisy (as in the SGD setting), when the model is provided with more substantial evidence, this additional noisy supervision can, at times, hurt generalization.
Therefore, enabling models to perform \textit{weight learning}, where the model adaptively weights the importance of symbolic rules as stronger evidence is introduced, is an interesting future direction \citep{karamanolakis2021self}.

%% file: sections/ack.tex
\section*{Acknowledgments}
This work was partially supported by the National Science Foundation grant CCF-2023495.

%% file: appendix/constraints.tex
\begin{figure*}[t] 
    \centering
    \noindent\fbox{%
        \begin{minipage}{0.99\hsize}
            \begin{scriptsize}
            \begin{flalign*}
                \hspace{0.2cm} & \textit{\# Token Constraint} && \\
                & w_{1}: \pslpred{HasWord}(\pslarg{Utt}, \pslarg{Class}) \rightarrow \pslpred{State}(\pslarg{Utt}, \pslarg{Class}) \\
            \end{flalign*}
            \end{scriptsize}
        \end{minipage}
    }
    \caption{SGD Structure Induction Constraint Model}
    \label{fig:sgd-constraints}
\end{figure*}

\section{Model Details}
\label{sec:model-details}

This section provides additional details on the \shortname{} models for the Multi-WoZ and SGD settings.
Throughout these subsections, we cover the symbolic constraints and the hyperparameters used.
All unspecified values for the constraints or the DD-VRNN model were left at their default values.
The code is under the Apache 2.0 license.

\subsection{SGD Constraints}
\label{appendix-model-details-sgd-constraints}

The \shortname \ model uses a single constraint for all SGD settings (synthetic, standard, domain generalization, and domain adaptation).
\figref{fig:sgd-constraints} provides an overview of the constraint, which contains the following two predicates:

\begin{itemize}
    \item[1.] \textbf{$\pslpred{State}(\pslarg{Utt}, \pslarg{Class})$}\\
    The $\pslpred{State}$ continuous valued predicate is the probability that an utterance, identified by the argument $\pslarg{Utt}$, belongs to a dialog state, identified by the argument $\pslarg{Class}$.
    For instance, the utterance $hello\ world\ !$ for the $greet$ dialog state would create a predicate with a value between zero and one, i.e., $\pslpred{State}(hello\ world\ ! greet) = 0.7$.
    \item[2.] \textbf{$\pslpred{HasWord}(\pslarg{Utt}, \pslarg{Class})$}\\
    The $\pslpred{HasWord}$ binary predicate indicates if an utterance, identified by the argument $\pslarg{Utt}$, contains a known token for a particular class, identified by the argument $\pslarg{Class}$.
    For instance if a known token associated with the $greet$ class is $hello$, then the utterance $hello\ world\ !$ would create a predicate with value one, i.e. $\pslpred{HasWord}(hello\ world\ !, greet) = 1$.
\end{itemize}

This token constraint encodes the prior knowledge that utterances' are likely to belong to dialog states when an utterance contains tokens representing that state.
For example, if a known token associated with the $greet$ class is $hello$, then the utterance $hello\ world \ !$ is likely to belong to the $greet$ state.
The primary purpose of incorporating this constraint into the model is to show how even a small amount of prior knowledge can aid predictions.
To get the set of tokens associated with each state, we trained a supervised linear classifier where the input is an utterance, and the label is the class.
After training, every token is individually run through the trained model to get a set of logits over each class.
These logits represent the relative importance that each token has over every class.
Sparsity is introduced to this set of logits, leaving only the top 0.1\% of values and replacing the others with zeros.
This sparsity reduces the set of 261,651 logits to 262 non-zero logits.

\begin{figure*}[t] 
    \centering
    \noindent\fbox{%
        \begin{minipage}{0.99\hsize}
            \begin{scriptsize}
            \begin{flalign*}
                \hspace{0.2cm} & \textit{\# Dialog Start} && \\
                % !FirstStatement -> !State('greet')
                & w_{1}: \pslneg \pslpred{FirstUtt}(\pslarg{Utt}) \rightarrow \ \pslneg \pslpred{State}(\pslarg{Utt}, greet) \\
                % FirstStatement(S) & HasGreetWord(S) -> State(S, 'greet')
                & w_{2}: \pslpred{FirstUtt}(\pslarg{Utt}) \psland \pslpred{HasGreetWord}(\pslarg{Utt}) \rightarrow \pslpred{State}(\pslarg{Utt}, greet) \\
                % FirstStatement(S) & !HasGreetWord(S) -> State(S, 'init_request')
                & w_{3}: \pslpred{FirstUtt}(\pslarg{Utt}) \psland \pslneg \pslpred{HasGreetWord}(\pslarg{Utt}) \rightarrow \pslpred{State}(\pslarg{Utt}, init\_request) \\[0.3cm]
                & \textit{\# Dialog Middle} \\
                % PreviousStatement(S1, S2) & State(S2, 'greet') -> State(S1, 'init_request')
                & w_{4}: \pslpred{PrevUtt}(\pslarg{Utt1}, \pslarg{Utt2}) \psland \pslpred{State}(\pslarg{Utt2}, greet) \rightarrow \pslpred{State}(\pslarg{Utt1}, init\_request) \\
                % PreviousStatement(S1, S2) & !State(S2, 'greet') -> !State(S1, 'init_request')
                & w_{5}: \pslpred{PrevUtt}(\pslarg{Utt1}, \pslarg{Utt2}) \psland \pslneg \pslpred{State}(\pslarg{Utt2}, greet) \rightarrow \pslneg \pslpred{State}(\pslarg{Utt1}, init\_request) \\
                % PreviousStatement(S1, S2) & State(S2, 'initial request') -> State(S1, 'second request')
                & w_{6}: \pslpred{PrevUtt}(\pslarg{Utt1}, \pslarg{Utt2}) \psland \pslpred{State}(\pslarg{Utt2}, init\_request) \rightarrow \pslpred{State}(\pslarg{Utt1}, second\_request) \\
                % PreviousStatement(S1, S2) & State(S2, 'second_request') & HasInfoQuestionWord(S1) -> State(S1, 'info_question')
                & w_{7}: \pslpred{PrevUtt}(\pslarg{Utt1}, \pslarg{Utt2}) \psland \pslpred{State}(\pslarg{Utt2}, second\_request) \psland \pslpred{HasInfoQuestionWord}(\pslarg{Utt1}) \rightarrow \pslpred{State}(\pslarg{Utt1}, info\_question) \\
                % PreviousStatement(S1, S2) & State(S2, 'second_request') & HasSlotQuestionWord(S1) -> State(S1, 'slot_question')
                & w_{8}: \pslpred{PrevUtt}(\pslarg{Utt1}, \pslarg{Utt2}) \psland \pslpred{State}(\pslarg{Utt2}, second\_request) \psland \pslpred{HasSlotQuestionWord}(\pslarg{Utt1}) \rightarrow \pslpred{State}(\pslarg{Utt1}, slot\_question) \\
                % PreviousStatement(S1, S2) & State(S2, 'end') & HasCancelWord(S1) -> State(S1, 'cancel')
                & w_{9}: \pslpred{PrevUtt}(\pslarg{Utt1}, \pslarg{Utt2}) \psland \pslpred{State}(\pslarg{Utt2}, end) \psland \pslpred{HasCancelWord}(\pslarg{Utt1}) \rightarrow \pslpred{State}(\pslarg{Utt1}, cancel) \\[0.3cm]
                & \textit{\# Dialog End} \\
                % LastStatement(S) & HasEndWord(S) -> State(S, 'end')
                & w_{10}: \pslpred{LastUtt}(\pslarg{Utt}) \psland \pslpred{HasEndWord}(\pslarg{Utt}) \rightarrow \pslpred{State}(\pslarg{Utt}, end) \\
                % LastStatement(S) & HasAcceptWord(S) -> State(S, 'accept')
                & w_{11}: \pslpred{LastUtt}(\pslarg{Utt}) \psland \pslpred{HasAcceptWord}(\pslarg{Utt}) \rightarrow \pslpred{State}(\pslarg{Utt}, accept) \\
                % LastStatement(S) & HasInsistWord(S) -> State(S, 'insist')
                & w_{12}: \pslpred{LastUtt}(\pslarg{Utt}) \psland \pslpred{HasInsistWord}(\pslarg{Utt}) \rightarrow \pslpred{State}(\pslarg{Utt}, insist) \\
            \end{flalign*}
            \end{scriptsize}
        \end{minipage}
    }
    \caption{MultiWoZ Structure Induction Constraint Model}
    \label{fig:multiwoz-constraints}
\end{figure*}

\subsection{Multi-WoZ Constraints}

The \shortname \ model for the Multi-WoZ setting uses a set of dialog constraints, which can be broken into dialog start, middle, and end.
\figref{fig:multiwoz-constraints} provides an overview of the constraints, which contains the following 11 predicates:

\begin{itemize}
    \item[1.] \textbf{$\pslpred{State}(\pslarg{Utt}, \pslarg{Class})$}\\
    The $\pslpred{State}$ continuous valued predicate is the probability that an utterance, identified by the argument $\pslarg{Utt}$, belongs to a dialog state, identified by the argument $\pslarg{Class}$.
    For instance, the utterance $hello\ world\ !$ for the $greet$ dialog state would create a predicate with a value between zero and one, i.e., $\pslpred{State}(hello\ world\ ! greet) = 0.7$.
    \item[2.]\textbf{$\pslpred{FirstUtt}(\pslarg{Utt})$}\\
    The $\pslpred{FirstUtt}$ binary predicate indicates if an utterance, identified by the argument $\pslarg{Utt}$, is the first utterance in a dialog.
    \item[3.] \textbf{$\pslpred{LastUtt}(\pslarg{Utt})$}\\
    The $\pslpred{LastUtt}$ binary predicate indicates if an utterance, identified by the argument $\pslarg{Utt}$, is the last utterance in a dialog.
    \item[4.] \textbf{$\pslpred{PrevUtt}(\pslarg{Utt1, Utt2})$}\\
    The $\pslpred{PrevUtt}$ binary predicate indicates if an utterance, identified by the argument $\pslarg{Utt2}$, is the previous utterance in a dialog of another utterance, identified by the argument $\pslarg{Utt1}$.
    \item[5.] \textbf{$\pslpred{HasGreetWord}(\pslarg{Utt})$}\\
    The $\pslpred{HasGreetWord}$ binary predicate indicates if an utterance, identified by the argument $\pslarg{Utt}$, contains a known token for the greet class.
    The list of known greet words is $['hello', 'hi']$.
    \item[6.] \textbf{$\pslpred{HasInfoQuestionWord}(\pslarg{Utt})$}\\
    The $\pslpred{HasInfoQuestionWord}$ binary predicate indicates if an utterance, identified by the argument $\pslarg{Utt}$, contains a known token for the info question class.
    The list of known info question words is $['address', 'phone']$.
    \item[7.] \textbf{$\pslpred{HasSlotQuestionWord}(\pslarg{Utt})$}\\
    The $\pslpred{HasSlotQuestionWord}$ binary predicate indicates if an utterance, identified by the argument $\pslarg{Utt}$, contains a known token for the slot question class.
    The list of known slot question words is $['what', '?']$.
    \item[8.] \textbf{$\pslpred{HasInsistWord}(\pslarg{Utt})$}\\
    The $\pslpred{HasInsistWord}$ binary predicate indicates if an utterance, identified by the argument $\pslarg{Utt}$, contains a known token for the insist class.
    The list of known insist words is $['sure', 'no']$.
    \item[9.] \textbf{$\pslpred{HasCancelWord}(\pslarg{Utt})$}\\
    The $\pslpred{HasCancelWord}$ binary predicate indicates if an utterance, identified by the argument $\pslarg{Utt}$, contains a known token for the cancel class.
    The list of known cancel words is $['no']$.
    \item[10.] \textbf{$\pslpred{HasAcceptWord}(\pslarg{Utt})$}\\
    The $\pslpred{HasAcceptWord}$ binary predicate indicates if an utterance, identified by the argument $\pslarg{Utt}$, contains a known token for the accept class.
    The list of known accept words is $['yes', 'great']$.
    \item[11.] \textbf{$\pslpred{HasEndWord}(\pslarg{Utt})$}\\
    The $\pslpred{HasEndWord}$ binary predicate indicates if an utterance, identified by the argument $\pslarg{Utt}$, contains a known token for the end class.
    The list of known end words is $['thank', 'thanks']$.
\end{itemize}

The dialog start constraints take advantage of the inherent structure built into the beginning of task-oriented dialogs.
In the same order as the dialog start rules in \figref{fig:multiwoz-constraints}:
1) If the first turn utterance does not contain a known greet word, then it does not belong to the $greet$ state.
2) If the first turn utterance contains a known greet word, then it belongs to the $greet$ state.
3) If the first turn utterance does not contain a known greet word, then it belongs to the $initial \ request$ state.

The dialog middle constraints exploit the temporal dependencies within the middle of a dialog.
In the same order as the dialog middle rules in \figref{fig:multiwoz-constraints}:
1) If the previous utterance belongs to the $greet$ state, then the current utterance belongs to the $initial \ request$ state.
2) If the previous utterance does not belong to the $greet$ state, then the current utterance does not belong to the $initial \ request$ state.
3) If the previous utterance belongs to the $initial \ request$ state, then the current utterance belongs to the $second \ request$ state.
4) If the previous utterance belongs to the $second \ request$ state and it has a known info question token, then the current utterance belongs to the $info \ question$ state.
5) If the previous utterance belongs to the $second \ request$ state and it has a known slot question token, then the current utterance belongs to the $slot \ question$ state.
4) If the previous utterance belongs to the $end$ state and it has a known cancel token, then the current utterance belongs to the $cancel$ state.

The dialog end constraints take advantage of the inherent structure built into the end of task-oriented dialogs.
In the same order as the dialog end rules in \figref{fig:multiwoz-constraints}:
1) If the last turn utterance contains a known end word, then it belongs to the $end$ state.
2) If the last turn utterance contains a known accept word, then it belongs to the $accept$ state.
3) If the last turn utterance contains a known insist word, then it belongs to the $insist$ state.

\begin{figure*}[t]
    \centering
    
    \includegraphics[width=1.00\textwidth]{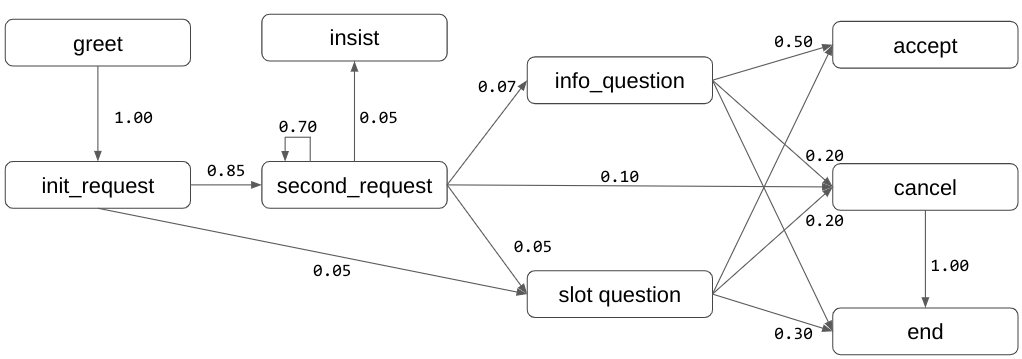}

    \caption{
        Ground truth dialog structure used to generate the MultiWoZ 2.1 dataset.
        The transition graph shows transitions over 0.05\%.
    }
    \label{fig:multiwoz_ground_truth}
\end{figure*}

\section{Additional Model Details}

\subsection{Symbolic-rule Normalization in the Multi-class Setting}
\label{appendix-normalization}
In the multi-class setting (e.g., multiple latent states), some soft logic operation on the model probability $p_\textbf{w}$ will lead to a probability that no longer normalizes to 1.
For example, the negation operation on the probability vector $p_\textbf{w}$ will lead to $!p_\textbf{w} = 1 - p_\textbf{w}$; then in the multi-class setting, the norm of $!p_\textbf{w}$ is $\sum_{i}^{|C|}(1 - p_i) = |C| - 1 > 1$, where $|C|$ is the number of classes. To address the above concern, we re-normalize after every soft logic operation:
\begin{align*}
    f_\mathbf{w}(\mathbf{y}, \mathbf{x}) = 
    f_\mathbf{w}(\mathbf{y}, \mathbf{x}) / ||f_\mathbf{w}&(\mathbf{y}, \mathbf{x})||,
\end{align*}
where $f_\mathbf{w} (\mathbf{y}, \mathbf{x})$ is the output of a soft logical operation.

\subsection{Model Hyperparameters}
\label{sec:model-details-hyperparameters}

The \textit{DD-VRNN} uses an LSTM \cite{hochreiter:nc97} with 200-400 units for the RNNs, and fully-connected highly flexible feature extraction functions with a dropout of 0.4 for the input x, the latent vector z, the prior, the encoder and the decoder.
The input to the \textit{DD-VRNN} is the utterances with a 300-dimension word embedding created using a GloVe embedding \cite{pennington:emnlp14} and a Bert embedding \citep{devlin-etal-2019-bert}.
The maximum utterance word length was set to 40, the maximum length of a dialog was set to 10, and the tunable weight, $\gamma$ (\eqnref{eqn:background-dd-vrnn-loss}), was set to 0.1. The total number of parameters is 26,033,659 for the model with GloVe embedding and 135,368,227 with Bert embedding.
The experiments are run in Google TPU V4, and the total GPU hours for all finetuning are 326 GPU hours.

%% file: appendix/datasets.tex
\section{Datasets}
\label{sec:data_app}

This section provides additional information on the SGD, SGD synthetic, and MultiWoZ 2.1 synthetic datasets.

\subsection{SGD}
 The Schema-Guided Dialog (SGD) \cite{rastogi2020towards} is a task-oriented conversation dataset involving interactions with services and APIs covering 20 domains.
 There are overlapping functionalities over many APIs, but their interfaces differ.
 One conversion may involve multiple domains.
 The train set contains conversions from 16 domains, with four held-out domains only present in test sets.
 This gives 34,308 in-domain and 5,441 out-of-domain test examples.
 To evaluate the model's generalization, we evaluate the model performance on both test sets.
 In specific, we establish three different evaluation protocols.
 \begin{itemize}
    \item \textbf{SGD Standard Generalization} We train the model using the SGD train set and evaluate it on the in-domain test set.  
    \item \textbf{SGD Domain Generalization}  We train the model using the SGD train set and evaluate it on the out-of-domain test set.
    \item \textbf{SGD Domain Adaptation}  We train the model using the SGD train set and label-wiped in-domain and out-of-domain test sets and evaluate it on the out-of-domain test set.
 \end{itemize}
\subsection{SGD Synthetic}
Using the template-based generator from the SGD developers \citet{kale-rastogi-2020-template}, we generate 10,800 synthetic dialogs using the same APIs and dialog states as the official SGD data. We split the examples with 75\% train and 25\% test. The schema-guided generator code is under Apache 2.0 license:  https://github.com/google-research/task-oriented-dialogue/blob/main/LICENSE.

\begin{table*}[t]
    \centering
    \resizebox{\textwidth}{!}{
        \begin{tabular}{cc|ccc|c|c}
            \toprule
                % \multirow{3}{*}{}
                % \multicolumn{3}{c}{}
                \multirow{2}{*}{Metric} & \multirow{2}{*}{Method} & \multicolumn{3}{c|}{SGD} & \multirow{2}{*}{SGD Synthetic} & \multirow{2}{*}{MultiWoZ} \\
                & & Standard & Domain Generalization & Domain Adaptation & & \\
             \midrule
                \multirow{3}{*}{Purity} & Random & 0.098 ± 0.000 & 0.098 ± 0.000 & 0.098 ± 0.000 & 0.094 ± 0.001 & 0.480 ± 0.000 \\
                & DD-VRNN & 0.341 ± 0.019 & 0.425 ± 0.016 & \textbf{0.443 ± 0.015} & 0.447 ± 0.024 & 0.701 ± 0.042 \\
                & \shortname & \textbf{0.463 ± 0.039} & \textbf{0.468 ± 0.039} & 0.425 ± 0.056 & \textbf{0.810 ± 0.005} & \textbf{0.762 ± 0.015} \\
            \bottomrule
        \end{tabular}
    }
    
    \caption{Test set performance on MultiWoZ Synthetic, SGD, and SGD Synthetic. These values correlate with the results reported in \tabref{tab:structure-learning-performance}.}
    \label{tab:structure-learning-performance-purity-class-balanced-accuracy}
\end{table*}

\subsection{MulitWoZ 2.1 Synthetic}

MultiWoZ 2.1 synthetic \cite{campagna:acl20} is a multi-domain goal-oriented dataset covering five domains (Attraction, Hotel, Restaurant, Taxi, and Train) and nine dialog acts ($greet$, $initial\ request$, $second\ request$, $insist$, $info\ question$, $slot\ question$, $accept$, $cancel$, and $end$).
Following \citenoun{campagna:acl20}, we generate $10^4$ synthetic dialogs from a known ground-truth dialog structure.
\figref{fig:multiwoz_ground_truth} provides an overview of the ground truth dialog structure, which is based on the original MultiWoz 2.1 dataset \cite{eric:lrec19}, used through the generative process.
These $10^4$ synthetic dialogs are randomly sampled without replacement to create ten splits with 80\% train, 10\% test, and 10\% validation.
The MultiWoZ 2.1 synthetic code is under the MIT License: https://github.com/stanford-oval/zero-shot-multiwoz-acl2020.
The MultiWoZ 2.1 code uses genie under the MIT License: https://github.com/stanford-oval/genie-k8s/blob/master/LICENSE.

%% file: appendix/extended_experiments.tex
\begin{figure*}[t]
    \centering
    
    \includegraphics[width=0.95\textwidth]{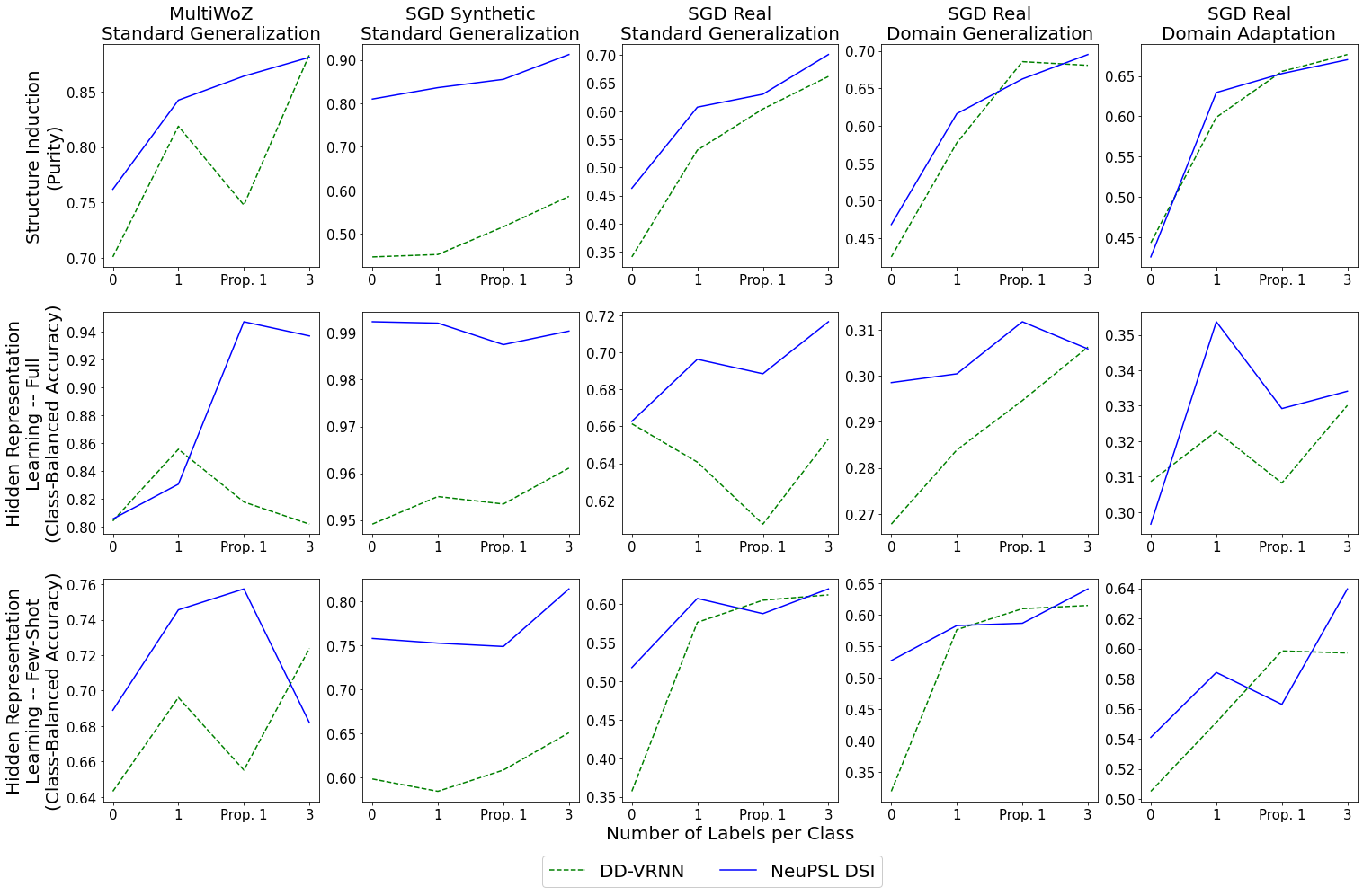}

    \caption{
        Average Purity and Class Balanced Accuracy on MultiWoZ Synthetic, SGD, and SGD Synthetic for varying amounts of supervision. These values correlate with the results reported in \figref{fig:main_results_few_shot}.
    }
    \label{fig:main_results_few_shot_extended}
\end{figure*}

\section{Extended Experimental Evaluation}

In this section, we provide extended experimental evaluations on the \shortname{} models for all settings.
We split the extended evaluation into evaluation metrics, main results, ablation results, and additional experiments.
Details describing changes to the models are provided in each subsection.

\subsection{Evaluation Metrics}
\label{sec:ami_app}

\paragraph{Adjusted Mutual Information (AMI) - }
AMI evaluates dialog structure prediction by evaluating the correctness of the dialog state assignments.
Let $U^*=\{U^*_1, \dots, U^*_{C^*}\}$ be the ground-truth assignment of dialog states for all utterances in the corpus, and $U=\{U_1, \dots, U_C\}$ be the predicted assignment of dialog states based on the learned dialog structure model.
$U^*$ and $U$ are not directly comparable because they draw from different base sets of states ($U*$ from the ground truth set of states and $U$ from the collection of states induced by the DD-VRNN) that may even have different cardinalities.
We address this problem using Adjusted Mutual Information (AMI), a metric developed initially to compare unsupervised clustering algorithms.
Intuitively, AMI treats each assignment as a probability distribution over states and uses Mutual Information to measure their similarity, adjusting for the fact that larger clusters tend to have higher MI.
AMI is defined as follows:
{
\begin{align*}
    AMI&(U, U^*) = \\
    & \dfrac{MI(U, U^*) - \mathbb{E}(MI(U, U^*))}{Avg(H(U), H(U^*)) - \mathbb{E}(MI(U, U^*))}
\end{align*}
} %
Where $MI(U, U^*)$ is the mutual information score, $\mathbb{E}(MI(U, U^*))$ is the expected mutual information over all possible assignments, and $Avg(H(U), H(U^*))$ is the average entropy of the two clusters \citep{vinh:jmlr10}.

\paragraph{Purity -}
Let $U^*=\{U^*_1, \dots, U^*_{C^*}\}$ be the ground-truth assignment of dialog states for all utterances in the corpus, and $U=\{U_1, \dots, U_C\}$ be the predicted assignment of dialog states based on the learned dialog structure model.
Each cluster is assigned to the class which is most frequent in the cluster.
This assignment then calculates accuracy by summing together the total correct of each cluster and dividing by the total number of clusters.
Purity is defined as follows:
\begin{align*}
    Purity&(U, U^*) = \frac{1}{N} \sum_{k=1}^{K}Count(U, U^*, A_{k})
\end{align*}
where $K$ is the number of unique clusters predicted, $N$ is the total number of predicted utterances, $A_{k}$ is the most frequent underlying ground truth in cluster $k$, and $Count(U, U^{*}, A_{k})$ is the total number of correctly labeled utterances within that assigned cluster.

\begin{table*}[ht!]
    \centering
    \resizebox{\textwidth}{!}{
        \begin{tabular}{cccc|cc|c}
            \toprule
                % \multirow{3}{*}{}
                % \multicolumn{3}{c}{}
                \multirow{3}{*}{\shortstack{Bag-of-Words\\Weights}}&\multirow{3}{*}{\shortstack{Constraint\\Loss}} & \multirow{3}{*}{\shortstack{Constraints\\Normalized}} & \multirow{3}{*}{Embedding} & \multicolumn{2}{c}{Hidden Representation Learning} & \multirow{2}{*}{Structure Induction}\\
                & & & & Full & Few-Shot & \\
                & & & & ( Class Balanced Accuracy ) & ( Class Balanced Accuracy )  & ( AMI )\\
                \midrule
                \multicolumn{4}{c|}{Random} & 0.111 ± 0.007 & 0.111 ± 0.007 & 0.000 ± 0.000 \\ \hline
                Uniform & Linear & Standard & Bert & \textbf{0.941 ± 0.010} & 0.667 ± 0.030 & 0.529 ± 0.040 \\
                Uniform & Linear & Standard & GloVe & 0.919 ± 0.015 & 0.672 ± 0.060 & 0.589 ± 0.050 \\
                Uniform & Linear & Normalized & Bert & \textbf{0.949 ± 0.008} & 0.645 ± 0.028 & 0.550 ± 0.018 \\
                Uniform & Linear & Normalized & GloVe & 0.934 ± 0.009 & \textbf{0.748 ± 0.057} & 0.516 ± 0.010 \\
                Uniform & Log & Standard & Bert & \textbf{0.944 ± 0.005} & 0.624 ± 0.039 & 0.586 ± 0.038 \\
                Uniform & Log & Standard & GloVe & 0.906 ± 0.008 & \textbf{0.711 ± 0.050} & 0.571 ± 0.011 \\
                Uniform & Log & Normalized & Bert & \textbf{0.944 ± 0.006} & \textbf{0.695 ± 0.027} & 0.505 ± 0.029 \\
                Uniform & Log & Normalized & GloVe & 0.918 ± 0.023 & 0.680 ± 0.057 & \textbf{0.612 ± 0.081} \\
                tf-idf & Linear & Standard & Bert & \textbf{0.943 ± 0.010} & 0.675 ± 0.035 & 0.574 ± 0.064 \\
                tf-idf & Linear & Standard & GloVe & 0.881 ± 0.016 & \textbf{0.744 ± 0.052} & \textbf{0.607 ± 0.061} \\
                tf-idf & Linear & Normalized & Bert & \textbf{0.947 ± 0.021} & \textbf{0.705 ± 0.021} & 0.511 ± 0.027 \\
                tf-idf & Linear & Normalized & GloVe & 0.925 ± 0.013 & \textbf{0.721 ± 0.051} & 0.544 ± 0.039 \\
                tf-idf & Log & Standard & Bert & \textbf{0.943 ± 0.007} & \textbf{0.705 ± 0.030} & 0.587 ± 0.027 \\
                tf-idf & Log & Standard & GloVe & 0.921 ± 0.016 & \textbf{0.747 ± 0.042} & \textbf{0.604 ± 0.012} \\
                tf-idf & Log & Normalized & Bert & \textbf{0.943 ± 0.005} & 0.689 ± 0.038 & \textbf{0.618 ± 0.028} \\
                tf-idf & Log & Normalized & GloVe & 0.913 ± 0.015 & \textbf{0.762 ± 0.070} & 0.545 ± 0.053 \\
            \bottomrule
        \end{tabular}
    }
    \caption{Test set performance on MultiWoZ Synthetic data setting.}
    \label{tab:ablation-multiwoz-synthetic}
\end{table*}

\subsection{Main Results}

\begin{table*}[ht!]
    \centering
    \resizebox{\textwidth}{!}{
        \begin{tabular}{cc|cc|c}
            \toprule
                \multirow{3}{*}{\shortstack{Bag-of-Words\\Weights}}&\multirow{3}{*}{\shortstack{Constraint\\Loss}} & \multicolumn{2}{c}{Hidden Representation Learning} & \multirow{2}{*}{Structure Induction}\\
                & & Full & Few-Shot & \\
                & & ( Class Balanced Accuracy ) & ( Class Balanced Accuracy )  & ( AMI )\\
             \midrule
                \multicolumn{2}{c|}{Random} & 0.026 ± 0.001 & 0.026 ± 0.001 & 0.000 ± 0.000 \\ \hline
                Uniform & Linear & 0.983 ± 0.003 & 0.717 ± 0.021 & 0.754 ± 0.032 \\
                Uniform & Log & \textbf{0.992 ± 0.003} & \textbf{0.758 ± 0.015} & 0.811 ± 0.005 \\
                Supervised & Linear & 0.988 ± 0.004 & 0.714 ± 0.021 & 0.746 ± 0.035 \\
                Supervised & Log & \textbf{0.993 ± 0.004} & 0.741 ± 0.019 & \textbf{0.820 ± 0.005} \\
            \bottomrule
        \end{tabular}
    }

    \caption{Test set performance on SGD Synthetic data setting.}
    \label{tab:ablation-sgd-synthetic}
\end{table*}

\begin{figure*}[!t]
    \centering

    \includegraphics[width=1.00\textwidth]{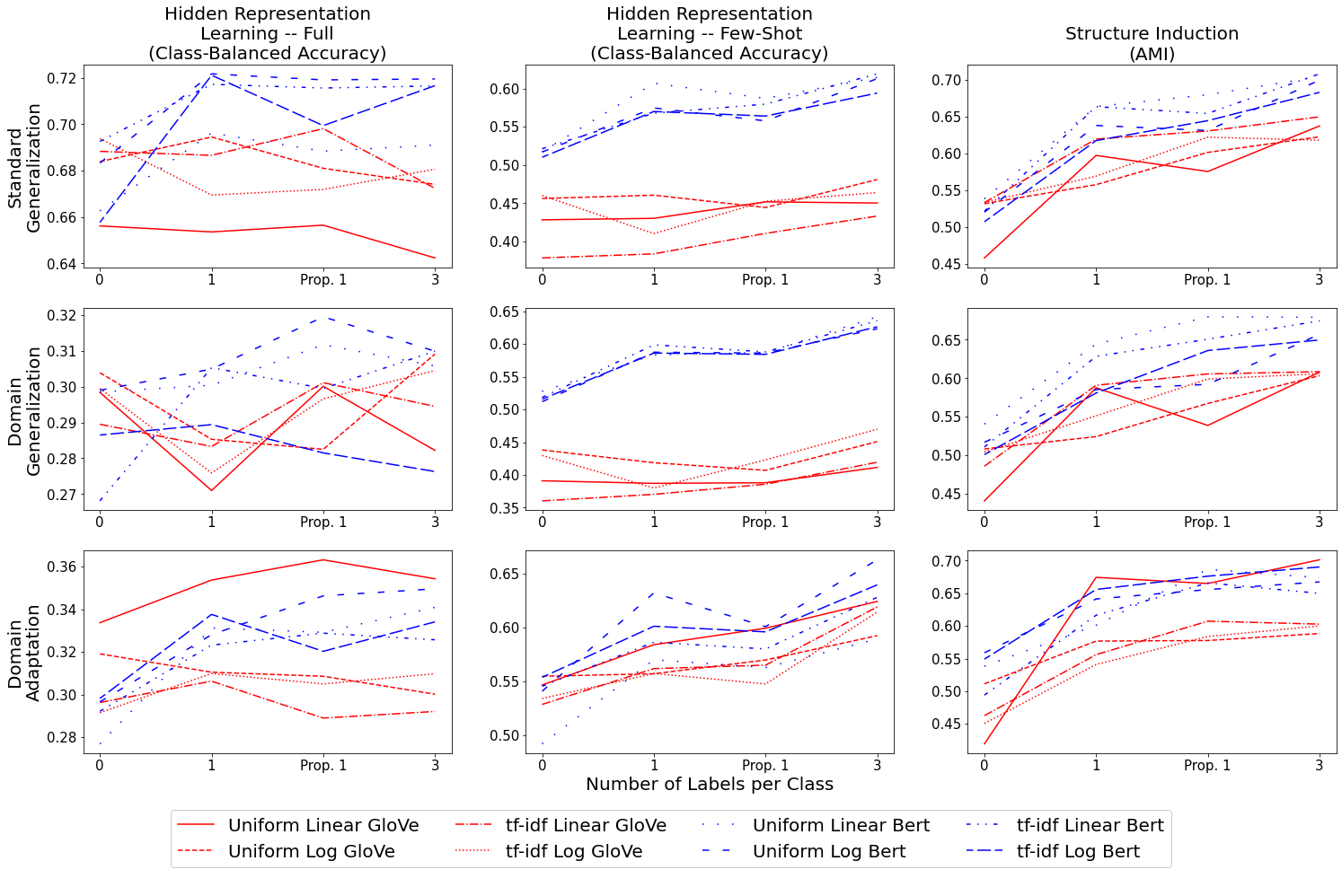}

    \caption{
        Average performance for SGD Real (Standard Generalization, Domain Generalization, and Domain Adaptation) on three highly constrained few-shot settings: 1-shot, proportional 1-shot, and 3-shot. Results are split into Hidden Representation Learning with class-balanced accuracy and Structure Induction with adjusted mutual information.
    }
    \label{fig:ablation_results_few_shot_sgd_real}
\end{figure*}

\begin{figure*}[!t]
    \centering
    
    \includegraphics[width=1.0\textwidth]{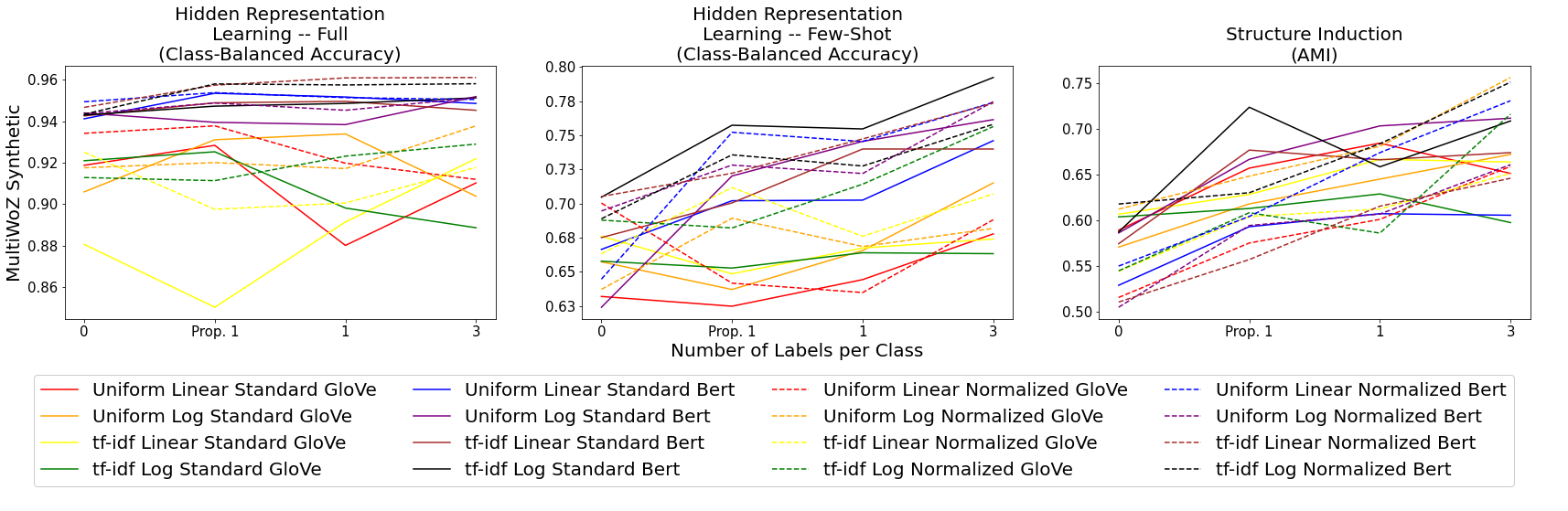}

    \caption{
        Average performance for MultiWoZ Synthetic on three highly constrained few-shot settings: 1-shot, proportional 1-shot, and 3-shot.
        Results are split into Hidden Representation Learning with class-balanced accuracy and Structure Induction with adjusted mutual information.
    }
    \label{fig:ablation_results_few_shot_multiwoz}
\end{figure*}

\begin{figure*}[!t]
    \centering
    
    \includegraphics[width=1.00\textwidth]{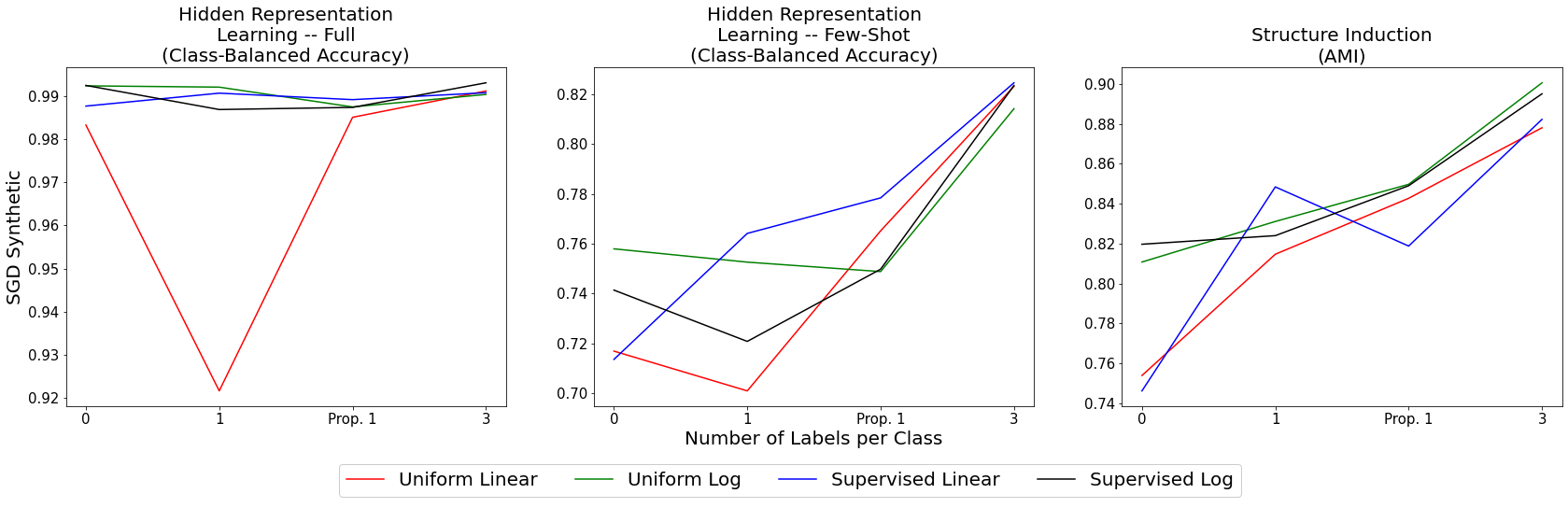}

    \caption{
        Average performance for SGD Synthetic on three highly constrained few-shot settings: 1-shot, proportional 1-shot, and 3-shot.
        Results are split into Hidden Representation Learning with class-balanced accuracy and Structure Induction with adjusted mutual information.
    }
    \label{fig:ablation_results_few_shot_synthetic}
\end{figure*}

This section provides additional experimental results for the structure induction and hidden representation learning performance.
\tabref{tab:structure-learning-performance-purity-class-balanced-accuracy} summarizes the extended evaluation of the main results for the \shortname{} model and DD-VRNN baseline on an additional metric: purity.
These values correlate with the reported results in \tabref{tab:structure-learning-performance}, i.e., these are not the best-performing results but are calculated using the model that produced the best AMI results.
Purity follows a similar trend as AMI, where \shortname{} outperforms the DD-VRNN in most settings.
In addition, \figref{fig:main_results_few_shot_extended} summarizes the few-shot results for purity and each hidden representation setting (full and few-shot).
Similar to the AMI, the introduction of labels improves performance across all settings.

\begin{table*}[t]
    \centering
    \resizebox{\textwidth}{!}{
        \begin{tabular}{cc|cc|c}
            \toprule
                Non-Zero & Non-Zero& \multicolumn{2}{c}{Hidden Representation Learning} & \multirow{2}{*}{Structure Induction}\\
                Word Weight & Word Weight & Full & Few-Shot & \\
                Percentage & Count & ( Class Balanced Accuracy ) & ( Class Balanced Accuracy )  & ( AMI )\\
             \midrule
                100.00\% & 261651 & 0.9997 ± 0.0006 & 0.9527 ± 0.0083 & 0.9999 ± 0.0001 \\
                3.25\% & 8499 & 0.9995 ± 0.0005 & 0.9636 ± 0.0028 & 0.9962 ± 0.0006 \\
                0.92\% & 2418 & 0.9995 ± 0.0002 & 0.9475 ± 0.0074 & 0.9616 ± 0.0010 \\
                0.42\% & 1111 & 0.9955 ± 0.0010 & 0.9213 ± 0.0053 & 0.9450 ± 0.0020 \\
                0.19\% & 504 & 0.9954 ± 0.0016 & 0.8591 ± 0.0082 & 0.7954 ± 0.0018 \\
                0.10\% & 262 & 0.9904 ± 0.0025 & 0.8241 ± 0.0243 & 0.8071 ± 0.0056 \\
                0.02\% & 54 & 0.9848 ± 0.0019 & 0.8193 ± 0.0111 & 0.6607 ± 0.0014 \\
                0.00\% & 0 & 0.9443 ± 0.0107 & 0.7283 ± 0.0127 & 0.5527 ± 0.0171 \\
            \bottomrule
        \end{tabular}
    }

    \caption{Test set performance on the SGD Synthetic data setting over varying sparsity in the token weights.}
    \label{tab:sparsity-results}
\end{table*}

\subsection{Ablation Results}
\label{sec:ablation_app}

This section provides an extended ablation for the SGD real setting and full ablations for the SGD synthetic and MultiWoZ datasets.
Each ablation analysis studies structure induction and hidden representation learning (\secref{sec:metrics_and_methodology}) over various neural settings.

\tabref{tab:ablation-multiwoz-synthetic} summarizes the unsupervised results for the MulitWoZ data setting over four major method axes: parameterization of the constraint loss (linear v.s. log constraint loss, Section \ref{sec:log-constraints}), weighting scheme for the bag-of-words loss (uniform v.s. tf-idf weights, Section \ref{sec:method_bow_reweighting}), constraint normalization (standard v.s. normalized, \secref{appendix-normalization}), and the choice of underlying utterance embedding (BERT \cite{devlin-etal-2019-bert} v.s. GloVe \cite{pennington:emnlp14}) leading to a total of $2^4=16$ settings.

\tabref{tab:ablation-sgd-synthetic} summarizes the unsupervised results for the SGD synthetic data setting over two major method axes: parameterization of the constraint loss (linear v.s. log constraint loss, Section \ref{sec:log-constraints}), and weighting scheme for the bag-of-words loss (uniform v.s. tf-idf weights, Section \ref{sec:method_bow_reweighting}) leading to a total of $2^2=4$ settings.

\figref{fig:ablation_results_few_shot_sgd_real}, \figref{fig:ablation_results_few_shot_multiwoz}, and \figref{fig:ablation_results_few_shot_synthetic} summarize the few-shot training results for the MultiWoZ, SGD synthetic, and SGD real (Standard Generalization, Domain Generalization, and Domain Adaptation) data settings when training with 1 shot, proportional 1 shot, and 3 shots.

\subsection{Additional Experiments}

Throughout this section, we provide additional dialog structure experiments to understand further when injecting common-sense knowledge as structural constraints is beneficial.
The additional experiments are broken into 1) A study of the sparsity introduced into the tokens in the SGD synthetic setting and 2) An exploration of an alternative soft logic formulation.

\begin{table*}[!t]
    \centering
    \resizebox{\textwidth}{!}{
        \begin{tabular}{ccc|cc|c}
            \toprule
                \multirow{3}{*}{Soft Logic} & \multirow{3}{*}{\shortstack{Bag-of-Words\\Weights}} & \multirow{3}{*}{\shortstack{Constraint\\ Loss}} & \multicolumn{2}{c}{Hidden Representation Learning} & \multirow{2}{*}{Structure Induction}\\
                & & & Full & Few-Shot & \\
                & & & ( Class Balanced Accuracy ) & ( Class Balanced Accuracy )  & ( AMI )\\
             \midrule
                \multicolumn{3}{c|}{Random} & 0.0261 ± 0.0013 & 0.0261 ± 0.0013 & 0.0000 ± 0.0004 \\ \hline
                \multirow{4}{*}{Lukasiewicz} & Uniform & Linear & 0.9188 ± 0.0150 & 0.6320 ± 0.0290 & 0.5892 ± 0.0496 \\
                & Uniform & Log & 0.9060 ± 0.0083 & 0.6574 ± 0.0184 & 0.5707 ± 0.0105 \\
                & tf-idf & Linear & 0.8807 ± 0.0164 & 0.6761 ± 0.0289 & 0.6066 ± 0.0605 \\
                & tf-idf & Log & 0.9210 ± 0.0160 & 0.6579 ± 0.0204 & 0.6037 ± 0.0120 \\
            \midrule
                \multirow{4}{*}{Product Real} & Uniform & Linear & 0.9151 ± 0.0566 & 0.6194 ± 0.0529 & 0.3928 ± 0.1881 \\
                & Uniform & Log & 0.8807 ± 0.0502 & 0.6174 ± 0.0525 & 0.4579 ± 0.1897 \\
                & tf-idf & Linear & 0.9176 ± 0.0369 & 0.6741 ± 0.0411 & 0.4392 ± 0.1903 \\
                & tf-idf & Log & 0.9232 ± 0.0147 & 0.6479 ± 0.0367 & 0.5202 ± 0.0455 \\
            \bottomrule
        \end{tabular}
    }
    
    \caption{Test set AMI and standard deviation on MulitWoZ data set on two soft logic relaxations.}
    \label{tab:ablation-multiwoz-ltn}
\end{table*}

\begin{figure*}[ht!]
    \centering
    
    \includegraphics[width=0.95\textwidth]{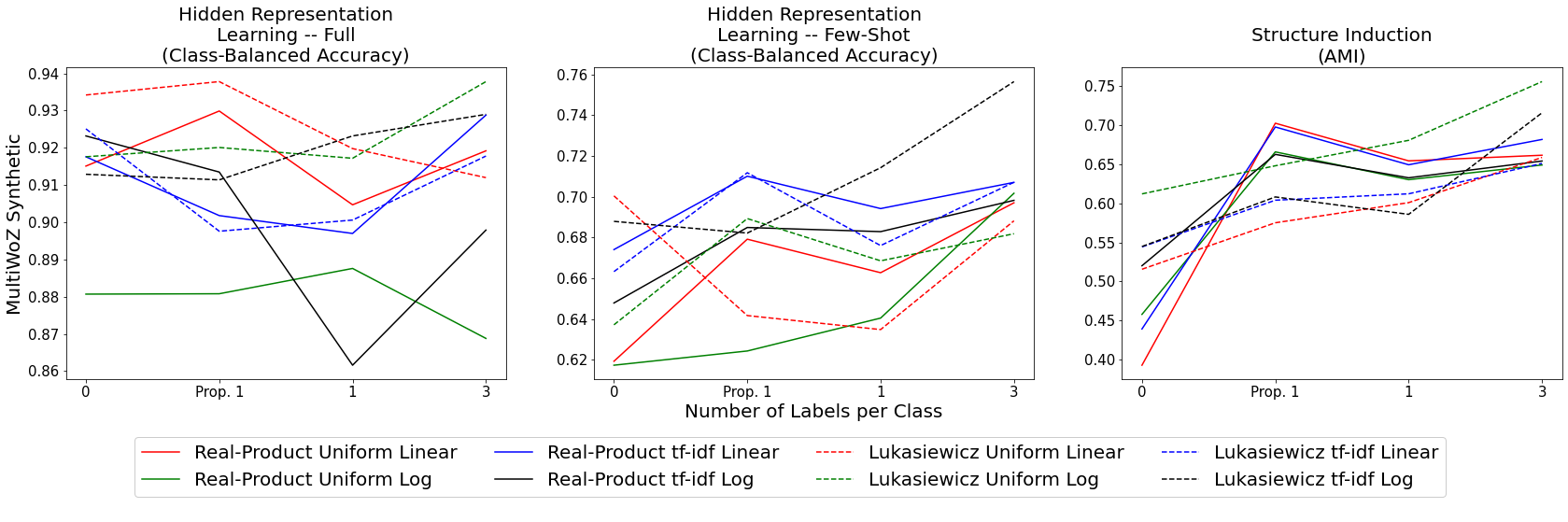}

    \caption{
        Average performance for MultiWoZ Synthetic for two soft logic relaxations (Real-Product and Lukasiewicz) on three highly constrained few-shot settings: 1-shot, proportional 1-shot, and 3-shot.
        Results are split into Hidden Representation Learning with class-balanced accuracy and Structure Induction with adjusted mutual information.
    }
    \label{fig:ablation_results_ltn_few_shot}
\end{figure*}

\subsubsection{Sparsity}

In this experiment, we explore varying the sparsity introduced to the token weights, as described in \appref{appendix-model-details-sgd-constraints}.
\tabref{tab:sparsity-results} shows the performance for the hidden representation and structure induction tasks.
When the percent of non-zero word weights is 100.00\%, this implies the model is trained on full supervision, while the non-zero word weights at 0.00\% represent the unsupervised DD-VRNN results.
Surprisingly, we see substantial improvement in all data settings.
Even when the non-zero word weight percentage is 0.02\%, resulting in 54 non-zero weights, we still see approximately a 20\% improvement to the AMI.
Note 54 non-zero weights are equivalent to about two identifiable tokens per class.

\section{Alternative Soft Logic Approximations}

This work is highly related to the active field of neuro-symbolic computing (NeSy) \citep{garcez:book02,bader:wwst05,garcez:book09,serafini:aiia16,besold:arxiv17,donadello:ijcai17,yang:neurips17,evans:jair18,manhaeve:ai21,garcez:jal19,deraedt:ijcai20,lamb:ijcai20,badreddine:ai22}.
NeSy methods aim to incorporate logic-based reasoning with neural networks.
As such, various principled soft/fuzzy logic formulations have shown to work quite well as knowledge used within the loss of a NeSy approach \citep{krieken:ai22, diligenti:ecml16, badreddine:ai22, manhaeve:ai21}.
While this work primarily focuses on using \textit{Lukasiewicz} logic \cite{klir:book95}, we are interested in how different soft/fuzzy logic formulations affect the learning process.
We, therefore, explore an alternative soft logic formulation, \textit{Product Real} logic, which is used in another principled NeSy framework called \textit{Logic Tensor Networks} \citep{badreddine:ai22}.
Similar to the \textit{Lukasiewicz} logic, Product Real logic approximates logical clauses with linear inequalities:
\begin{align*}
    A \psland B &= A * B\\
    A \pslor B &= A + B - A * B\\
    \pslneg A &= 1.0 - A
\end{align*}
where $A$ and $B$ are either ground atoms or logical expressions over atoms.
In either case, they have values between [0,1].

\tabref{tab:ablation-multiwoz-ltn} summarizes the unsupervised results for the MulitWoZ data setting over three major method axes: the soft logic approximation (Lukasiewicz v.s. Product Real), parameterization of the constraint loss (linear v.s. log constraint loss, Section \ref{sec:log-constraints}), and weighting scheme for the bag-of-words loss (uniform v.s. tf-idf weights, Section \ref{sec:method_bow_reweighting}) leading to a total of $2^3=8$ settings.
Surprisingly, in structure induction, Lukasiewicz logic outperformed Product Real logic by over 7\%.
Though interestingly, the hidden representation learning performance was roughly equivalent between the two soft logic formulations.
These findings highlight the importance of selecting the appropriate soft/fuzzy logic formulation for a given problem.
Further investigation into different soft logic formulations would be an interesting direction for future research to gain a more comprehensive understanding of their potential applications.

\figref{fig:ablation_results_ltn_few_shot} summarizes the few-shot training results for the MultiWoZ synthetic data settings when training with 1 shot, proportional 1 shot, and 3 shots.
Noticeably, the Product Real logic closes the gap with the introduction of labels.
However, the Lukasiewicz logic still has an edge when observing the largest semi-supervised setting.